\PassOptionsToPackage{unicode}{hyperref}
\PassOptionsToPackage{hyphens}{url}
\documentclass[
]{article}

\usepackage{arxiv}
\usepackage{amsmath,amssymb}
\usepackage{lmodern}
\usepackage{iftex}
\ifPDFTeX
  \usepackage[T1]{fontenc}
  \usepackage[utf8]{inputenc}
  \usepackage{textcomp} 
\else 
  \usepackage{unicode-math}
  \defaultfontfeatures{Scale=MatchLowercase}
  \defaultfontfeatures[\rmfamily]{Ligatures=TeX,Scale=1}
\fi
\IfFileExists{upquote.sty}{\usepackage{upquote}}{}
\IfFileExists{microtype.sty}{
  \usepackage[]{microtype}
  \UseMicrotypeSet[protrusion]{basicmath} 
}{}
\makeatletter
\@ifundefined{KOMAClassName}{
  \IfFileExists{parskip.sty}{%
    \usepackage{parskip}
  }{
    \setlength{\parindent}{0pt}
    \setlength{\parskip}{6pt plus 2pt minus 1pt}}
}{
  \KOMAoptions{parskip=half}}
\makeatother
\usepackage{xcolor}
\IfFileExists{xurl.sty}{\usepackage{xurl}}{} 
\IfFileExists{bookmark.sty}{\usepackage{bookmark}}{\usepackage{hyperref}}
\hypersetup{
  pdftitle={PosePipe: Open-Source Human Pose Estimation Pipeline for Clinical Research},
  pdfkeywords={Human Pose Estimation, DataJoint, Rehabilitation,
Movement Science},
  hidelinks,
  pdfcreator={LaTeX via pandoc}}
\usepackage{color}
\usepackage{fancyvrb}

\DefineVerbatimEnvironment{Highlighting}{Verbatim}{commandchars=\\\{\}}
\newenvironment{Shaded}{}{}
\newcommand{\AlertTok}[1]{\textcolor[rgb]{1.00,0.00,0.00}{\textbf{#1}}}

\newcommand{\BuiltInTok}[1]{#1}

\newcommand{\CommentTok}[1]{\textcolor[rgb]{0.38,0.63,0.69}{\textit{#1}}}

\newcommand{\ControlFlowTok}[1]{\textcolor[rgb]{0.00,0.44,0.13}{\textbf{#1}}}

\newcommand{\ImportTok}[1]{#1}

\newcommand{\KeywordTok}[1]{\textcolor[rgb]{0.00,0.44,0.13}{\textbf{#1}}}
\newcommand{\NormalTok}[1]{#1}
\newcommand{\OperatorTok}[1]{\textcolor[rgb]{0.40,0.40,0.40}{#1}}

\newcommand{\PreprocessorTok}[1]{\textcolor[rgb]{0.74,0.48,0.00}{#1}}

\newcommand{\SpecialCharTok}[1]{\textcolor[rgb]{0.25,0.44,0.63}{#1}}
\newcommand{\SpecialStringTok}[1]{\textcolor[rgb]{0.73,0.40,0.53}{#1}}
\newcommand{\StringTok}[1]{\textcolor[rgb]{0.25,0.44,0.63}{#1}}
\newcommand{\VariableTok}[1]{\textcolor[rgb]{0.10,0.09,0.49}{#1}}

\usepackage{graphicx}
\makeatletter
\def\maxwidth{\ifdim\Gin@nat@width>\linewidth\linewidth\else\Gin@nat@width\fi}
\def\maxheight{\ifdim\Gin@nat@height>\textheight\textheight\else\Gin@nat@height\fi}
\makeatother
\setkeys{Gin}{width=\maxwidth,height=\maxheight,keepaspectratio}
\makeatletter
\def\fps@figure{htbp}
\makeatother
\providecommand{\tightlist}{%
  \setlength{\itemsep}{0pt}\setlength{\parskip}{0pt}}
\newlength{\cslhangindent}
\newlength{\csllabelwidth}
\newlength{\cslentryspacingunit} 
\newenvironment{CSLReferences}[2] 
 {
  \setlength{\parindent}{0pt}
  \ifodd #1
  \let\oldpar\par
  \def\par{\hangindent=\cslhangindent\oldpar}
  \fi
  \setlength{\parskip}{#2\cslentryspacingunit}
 }%
 {}
\usepackage{calc}

\newcommand{\CSLLeftMargin}[1]{\parbox[t]{\csllabelwidth}{#1}}
\newcommand{\CSLRightInline}[1]{\parbox[t]{\linewidth - \csllabelwidth}{#1}\break}

\makeatletter
\@ifpackageloaded{subfig}{}{\usepackage{subfig}}
\@ifpackageloaded{caption}{}{\usepackage{caption}}
\AtBeginDocument{%

}
\AtBeginDocument{%

}
\newcounter{pandoccrossref@subfigures@footnote@counter}
{\end{figure}%
\addtocounter{footnote}{-\value{pandoccrossref@subfigures@footnote@counter}}
\@for\f:=\global@pandoccrossref@subfigures@footnotes\do{\stepcounter{footnote}\footnotetext{\f}}%
\gdef\global@pandoccrossref@subfigures@footnotes{}}
\@ifpackageloaded{float}{}{\usepackage{float}}
\floatstyle{ruled}
\@ifundefined{c@chapter}{\newfloat{codelisting}{h}{lop}}{\newfloat{codelisting}{h}{lop}[chapter]}
\floatname{codelisting}{Listing}

\makeatother
\ifLuaTeX
  \usepackage{selnolig}  
\fi

\usepackage{authblk}

\newbox{\myorcidaffilbox}
\sbox{\myorcidaffilbox}{\large\includegraphics[height=1.7ex]{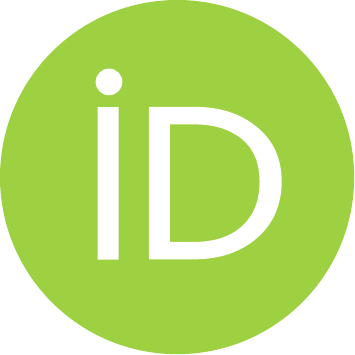}}
\newcommand{\orcidaffil}[1]{%
  \href{https://orcid.org/#1}{\usebox{\myorcidaffilbox}}}

\title{PosePipe: Open-Source Human Pose Estimation Pipeline for Clinical
Research}
\date{}
\author[*,1,2]{\orcidaffil{0000-0001-5714-1400} R. James Cotton}
\affil[*]{rcotton@sralab.org }
\affil[1]{Department of Physical Medicine and Rehabilitation, Northwestern University}
\affil[2]{Shirley Ryan AbilityLab}

\renewcommand{\headeright}{}
\renewcommand{\undertitle}{}

\begin{document}
\maketitle
\begin{abstract}
There has been significant progress in machine learning algorithms for
human pose estimation that may provide immense value in rehabilitation
and movement sciences. However, there remain several challenges to
routine use of these tools for clinical practice and translational
research, including: 1) high technical barrier to entry, 2) rapidly
evolving space of algorithms, 3) challenging algorithmic
interdependencies, and 4) complex data management requirements between
these components. To mitigate these barriers, we developed a human pose
estimation pipeline that facilitates running state-of-the-art algorithms
on data acquired in clinical context. Our system allows for running
different implementations of several classes of algorithms and handles
their interdependencies easily. These algorithm classes include subject
identification and tracking, 2D keypoint detection, 3D joint location
estimation, and estimating the pose of body models. The system uses a
database to manage videos, intermediate analyses, and data for
computations at each stage. It also provides tools for data
visualization, including generating video overlays that also obscure
faces to enhance privacy. Our goal in this work is not to train new
algorithms, but to advance the use of cutting-edge human pose estimation
algorithms for clinical and translation research. We show that this tool
facilitates analyzing large numbers of videos of human movement ranging
from gait laboratories analyses, to clinic and therapy visits, to people
in the community. We also highlight limitations of these algorithms when
applied to clinical populations in a rehabilitation setting. Code for
PosePipe can be found at
\url{https://github.com/peabody124/PosePipeline/}.
\end{abstract}

\hypertarget{introduction}{%
\section{Introduction}\label{introduction}}

Accurate tracking of human movement is a critical prerequisite for
movement science and rehabilitation research. The gold standard is a
movement analysis lab where optical markers are tracked with high
spatial and temporal precision in order to precisely reconstruct
biomechanical movements. While this method is highly accurate, it
requires significant expertise, is time consuming and expensive, and can
only be performed in a lab with specialized equipment. Wearable sensors
equipped with inertial measurement units can help track movement outside
of the laboratory, but often require significant time for setup and
calibration and are typically less accurate, as reconstructing the
underlying biomechanics from sensor data is a challenging data
processing
problem\textsuperscript{\protect\hyperlink{ref-poitras_validity_2019}{1}--\protect\hyperlink{ref-Filippeschi2017}{3}}.

Deep learning-based approaches to human pose estimation (HPE) from video
have advanced rapidly in recent
years\textsuperscript{\protect\hyperlink{ref-Zheng2020}{4}} and show
promise in enabling easy-to-use and precise movement analysis outside of
a specialized laboratory setting. These approaches could be a key
enabling technology for movement science and rehabilitation research.
For example, it could enable more frequent and precise measurements of
patient's movement during recovery and enable longitudinal
quantification of their movement impairments. This could allow better
understanding of how movement impairments relate to functional abilities
(i.e., ability to performing activities of daily living) and enable more
sensitive clinical trials to improve them. A recent consensus paper on
upper extremity rehabilitation after stroke highlighted the need for
more routine kinematic measurements for this express purpose, while also
pointing out this is impeded by the lack of easy to use measurement
tools\textsuperscript{\protect\hyperlink{ref-Kwakkel2019a}{5}}.

Despite recent advancements, there are still numerous barriers that
prevent video-based human pose estimation from fulfilling this goal for
rehabilitation and movement
science\textsuperscript{\protect\hyperlink{ref-Seethapathi2019}{6}}.
Among these barriers are: 1) there is a high technical barrier to using
many state-of-the-art algorithms; 2) these algorithms are rapidly
evolving; 3) many of the algorithms have interdependencies; 4) they
require a significant amount of data management; 5) the accuracy of many
of these algorithms have not been validated on clinical
populations\textsuperscript{\protect\hyperlink{ref-Parks2019}{7},\protect\hyperlink{ref-Needham2021}{8}};
and 6) many algorithms do not produce clinically pertinent outputs.
Here, we will briefly review some of the classes of algorithms for HPE
before demonstrating how our work on PosePipe reduces barriers 1-4
listed above. We also highlight some issues that occur when these
algorithms are tested on clinical populations, which relates to barrier
5, but we defer quantitative analyses of this and work on barrier 6 to
future work.

\hypertarget{types-of-hpe-algorithms}{%
\subsection{Types of HPE Algorithms}\label{types-of-hpe-algorithms}}

The field of human pose estimation from computer vision is large and
growing rapidly, and we refer interested readers to recent in-depth
reviews on the topic for additional
details\textsuperscript{\protect\hyperlink{ref-Zheng2020}{4},\protect\hyperlink{ref-Liu2021_Pose}{9}--\protect\hyperlink{ref-tian_recovering_2022}{11}}.
Here, we will provide a brief high-level overview of some common classes
of algorithms and taxonomic categories used in our framework and how
these interact.

HPE algorithms either process all people in a frame (bottom-up
approaches), or analyze a single person (top-down approaches). One
bottom-up approach utilized in numerous clinical studies
(e.g.\textsuperscript{\protect\hyperlink{ref-sato_quantifying_2019}{12}--\protect\hyperlink{ref-mehdizadeh_concurrent_2021}{15}})
is OpenPose\textsuperscript{\protect\hyperlink{ref-Cao2016}{16}}.
OpenPose locates keypoints in the image (e.g.~joint locations, finger
keypoints and some facial keypoints) for all people in the scene, and
groups them by individuals. However, if multiple people are visible in a
scene, then additional work is required to select the subject of
interest. In contrast, top-down approaches require the person of
interest to already be localized in each frame. For analyzing movement,
it is critical to consistently and accurately localize that particular
person throughout the video. Thus, the first step is running a tracking
algorithm, and if there are multiple people, selecting the subject for
analysis with subsequent algorithms. An additional benefit of top-down
approaches are that the keypoint accuracy tends to exceed bottom-up
approaches.

Top-down algorithms include a range of approaches with outputs that are
either two- or three-dimensional. Two-dimensional algorithms localize a
number of keypoints, such as the location of major joints in the image.
A strength of 2D keypoint detection is that the algorithms are fairly
mature and robust. However, to the best of our knowledge, there are no
large-scale systematic evaluations of this robustness in clinical
contexts or on patient populations. Therefore, it remains to be seen how
well they will generalize to these situations.

Three-dimensional approaches include those that predict 3D joint
locations, of which there are two common approaches. The first approach
is ``lifting'' 2D keypoints into 3D
coordinates\textsuperscript{\protect\hyperlink{ref-Martinez2017}{17},\protect\hyperlink{ref-Pavllo2018}{18}}.
Lifting algorithms are trained on datasets of paired 2D and 3D data and
use the implicit prior distribution over body configurations learned
from training data to resolve the inherent ambiguities involved in going
from 2D to 3D. The second approach estimates the parameters of a body
model, such as the Skinned Multi-Person Linear Model (SMPL)
model\textsuperscript{\protect\hyperlink{ref-Loper2015}{19}}, including
body shape and pose. Frequently, a neural network is trained to directly
predict these parameters from an image or
video\textsuperscript{\protect\hyperlink{ref-kanazawa_end_2018}{20}--\protect\hyperlink{ref-kocabas_vibe_2020}{22}}.
While these approaches provide a rich description of the joint angles
and body shape, 3D accuracy often lags behind lifting approaches
(e.g.\textsuperscript{\protect\hyperlink{ref-liu_graph_2020}{23}}).
Optimization based approaches can refine model fitting, but are much
slower than regression
approaches\textsuperscript{\protect\hyperlink{ref-Bogo2016}{24},\protect\hyperlink{ref-pavlakos_expressive_2019}{25}}.
The representation of pose is less sensitive to the perspective the
video was recorded from in 3D approaches, which provides a unique
advantage. However, none of these 3D approaches produce the joint angle
representations recommended by the International Standard of
Biomechanics\textsuperscript{\protect\hyperlink{ref-wu_isb_2002}{26},\protect\hyperlink{ref-Wu2005}{27}},
although we have previously shown that the SMPL parameters for the arm
can be converted to this
format\textsuperscript{\protect\hyperlink{ref-Cotton2020}{28}}.

\hypertarget{overview-of-our-approach}{%
\subsection{Overview of our approach}\label{overview-of-our-approach}}

Utilizing the evolving landscape of HPE tools, particularly when
managing large numbers of videos, presents several barriers. The first
is managing the dependencies between algorithms. For example, the
bounding boxes from tracking algorithms are used in top-down 2D keypoint
detection. The 2D keypoints sequences over time from a person are then
used by lifting algorithms to generate 3D keypoint trajectories. There
is no standardization over the formats of these datatypes, so typically
additional processing or reformatting is required to use the output of
one algorithm as input to another. A second challenge is managing data
from different processing stages and running them through a pipeline
when analyzing large sets of videos. Further, for each class of
algorithm, there are numerous different versions that have been released
with limited validation in clinical populations, making it hard \emph{a
priori} to determine the optimal combination of algorithms for a
particular question. Thus, it is important to have flexible pipelines
that allow analyzing videos using different algorithms, a task which is
made additionally difficult due to the lack of consistency in data
formats.

To facilitate HPE use in rehabilitation research and movement science,
we have developed an open-source tool that addresses and minimizes these
barriers. Video processing pipelines are created using
DataJoint\textsuperscript{\protect\hyperlink{ref-yatsenko_datajoint_2015}{29}},
which builds computational pipelines by managing all videos and outputs
in a MySQL database, and manages dependencies between computations. We
wrote wrappers for specific implementations of the algorithm classes
previously described (bounding box tracking, 2d keypoints, lifting, and
SMPL) that store data in DataJoint using a standardized format for each
step This standardization makes it easier to create a pipeline using any
particular set of implementations. Using newly developed algorithms on
data is also simplified in our method, as it only requires implementing
a new wrapper that uses the consistent format, rather than creating a
complete pipeline from scratch. This approach also allows mixing
algorithms implemented in different frameworks (e.g.~Jax, TensorFlow and
PyTorch). DataJoint also provides a job management system that allows
parallel computation across multiple GPUs or multiple computers with
minimal overhead and no code changes. Finally, PosePipe makes it easy to
visualize the outputs, which is essential when testing HPE algorithms on
clinical populations. We have used this pipeline to analyze 10s of
thousands of videos acquired in clinical settings, something only
possible due to PosePipe.

Having HPE results in DataJoint provides additional benefits, as it is a
very effective general data analysis tool with wide adoption in the
neuroscience
community\textsuperscript{\protect\hyperlink{ref-yatsenko_datajoint_2018}{30},\protect\hyperlink{ref-datajoint_team_datajoint_2022}{31}}.
By associating videos with experiment-specific, DataJoint schemas,
subsequent analysis becomes much easier (e.g., plotting longitudinal
summary statistics from subjects can be done with only a few lines). A
central database is also beneficial because multiple researchers on
separate computers can easily collaborate and access the same data with
a consistent organization and processing system. Code for PosePipe can
be found at \url{https://github.com/peabody124/PosePipeline/}.

\hypertarget{contributions}{%
\subsection{Contributions}\label{contributions}}

In short, our contributions in this work are:

\begin{itemize}
\tightlist
\item
  We describe and release PosePipe, a DataJoint based pipeline for HPE
  that facilitates large scale analysis of videos acquired in a clinical
  context
\item
  We provide wrappers to several state-of-the-art algorithms for HPE
  including for bounding box tracking, 2D keypoint estimation, 3D
  lifting, and estimating SMPL meshes. Implementing wrappers for newly
  released algorithms is substantially easier than creating custom
  pipelines to test algorithms.
\item
  We describe the pros and cons of different algorithm types and
  implementations and give examples of the types of errors that occur
  when using these algorithms on clinical populations.
\end{itemize}

\hypertarget{methods}{%
\section{Methods}\label{methods}}

PosePipe is written in Python and uses
DataJoint\textsuperscript{\protect\hyperlink{ref-yatsenko_datajoint_2015}{29}}
for data and computation management. We briefly highlight several
pertinent details of DataJoint that are important for understanding
PosePipe, but refer to the
\href{https://docs.datajoint.org/python/}{documentation} for details.
Under the DataJoint model, Python classes that inherit from DataJoint
base classes have a corresponding table in a database. DataJoint classes
can be of several types including: \texttt{Manual}, \texttt{Lookup}, and
\texttt{Computed}. As names suggest, \texttt{Manual} corresponds to rows
manually entered into the database (e.g.~by uploading videos),
\texttt{Lookup} corresponds to a lookup table and is commonly used to
indicate specific computation types, and \texttt{Computed} are tables
computed based on existing parent data in the database. When required
inputs are available, dependent rows in the database are automatically
computed (or populated) with the \texttt{populate} method on a
\texttt{Computed} table, which can also be performed for many entries in
parallel using the job management system built into DataJoint.

The relationship between the classes (i.e., tables) must correspond to a
directed acyclic diagram (DAG) and is described in the Python class
definitions and is enforced in the database through foreign key
constraints. This ensures data integrity; for example, a row can only
exist if the required rows in parent tables also exist. Individual rows
are identified in the database by primary keys, which are a set of
fields that must be unique. Child tables inherit the primary keys of
their parents (with the foreign key constraint) and can extend the
primary key with additional fields, which allows multiple descendent
rows (e.g., when using different specific algorithms or analyzing
multiple subjects of interest in a video).

\begin{figure}
\hypertarget{fig:schema}{%
\centering
\includegraphics[width=6.25in,height=\textheight]{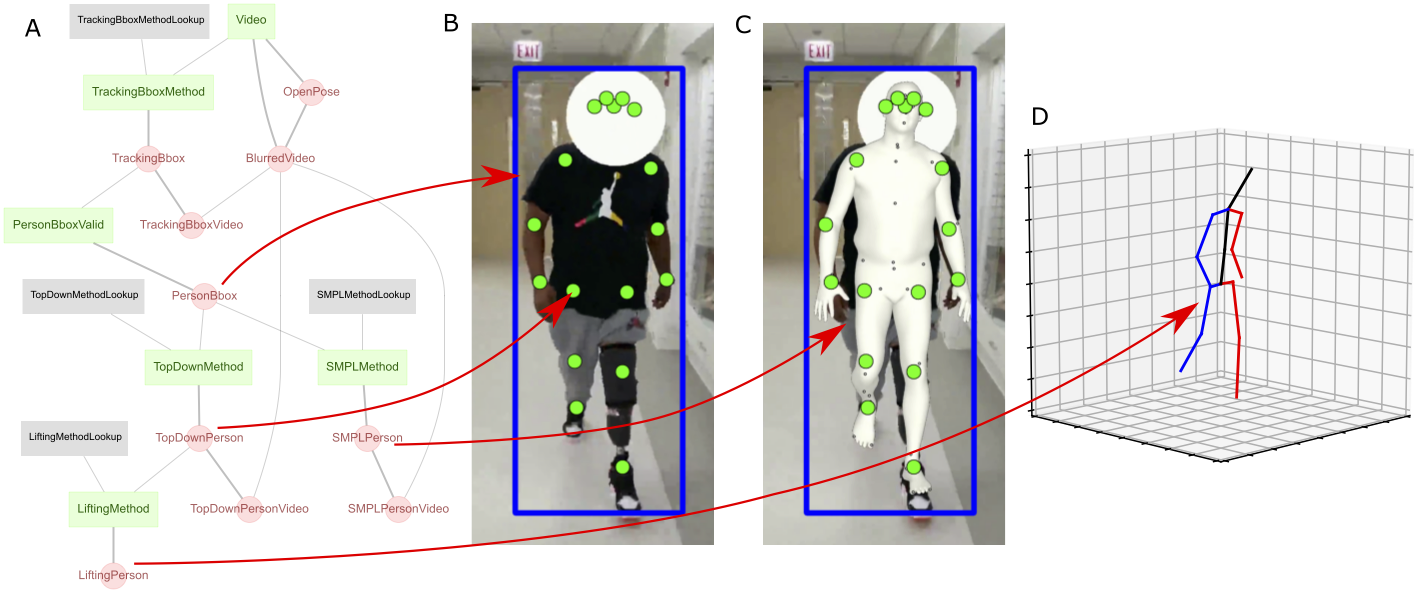}
\caption{A) Diagram of PosePipe computational tree, where each node
corresponds to a DataJoint class and table in the database. Green
squares correspond to manually entered rows, or rows imported by scripts
outside the pipeline. Grey squares indicate lookup tables enumerating
available methods. Circles indicate computed nodes, based on the
information in parent nodes. The lookup tables allow selecting specific
algorithms for each of these nodes. Arrows show the relationship between
the output nodes and example visualizations. B) Visualization showing
bounding box and 2D keypoints. C) Example SMPL mesh output. D) 3D
keypoint skeleton.}\label{fig:schema}
}
\end{figure}

The core classes (tables) for PosePipe and dependencies are diagramed in
Fig.~\ref{fig:schema}, which illustrates the structure previously
described, such as 2D keypoints depending on bounding box calculations
and 3D lifting keypoints depending on 2D keypoints. We also illustrate
examples of these outputs.

\hypertarget{posepipe-stages}{%
\subsubsection{PosePipe Stages}\label{posepipe-stages}}

\hypertarget{video-importing}{%
\paragraph{Video importing}\label{video-importing}}

Rows in the \texttt{Video} table Fig.~\ref{fig:schema} correspond to
individual videos imported into PosePipe. PosePipe provides optional
scripts to recompress videos with a consistent codec to ensure that
downstream analyses can read them. The primary keys for videos are two
strings: one for the filename and one for a project name. The latter
prevents potential name collisions for identically named videos from two
projects, and also makes it easy to restrict analysis to videos from a
particular project.

Because storing large files such as videos in a database can result in
poor performance, we use the DataJoint \texttt{attach} type to store the
videos externally on the filesystem. When retrieving a video from the
database, it is copied to the current working directory and the checksum
is validated against the original video checksum to ensure data
integrity. This design enables additional, more granular, access control
to the raw videos by controlling which users have filesystem permissions
to access the external video storage. This is a necessary and important
feature when working with clinical data where wider access might be
granted to the extracted movement trajectories stored in the database
compared to the raw videos that contain identifiable information, such
as faces. In a situation where multiple computers need to access the raw
videos, it does introduce an additional step of ensuring the same
directory is shared and available on all of the computers.

\hypertarget{subject-of-interest-identification}{%
\paragraph{Subject of Interest
Identification}\label{subject-of-interest-identification}}

Frequently, multiple people are visible in videos (i.e., patient,
physical therapist, and background individuals), but it is only
necessary to perform HPE on one or some of them. The most common
solution to this challenge are algorithms which first identify all
individuals in a frame by computing a bounding box that surrounds the
individual, followed by grouping the bounding boxes over time into
tracklets. These tracklets often cannot be used immediately. For
example, if there are multiple tracklets (for different people), it is
first necessary to identify which corresponds to the person of interest.
In some cases, the tracklets may also have two types of problems:
splitting or swapping.

In the case of splitting, the subject of interest is represented by
multiple tracklets at different points of times, sometimes with a gap.
This can occur because the subject was briefly occluded and the
algorithm failed to reidentify them when they reappeared. In other
cases, it can occur spontaneously due to a failure of the algorithm to
detect people. These separate tracklets can be manually linked to allow
consistent tracking throughtout the video, provided any gaps are not too
long. Swapping indicates that a single tracklet contains two different
people at different time points. In this case the tracklet cannot be
used without contaminating subsequent analyses. If the video contains
additional uncontaminated tracklets at different time points, these can
be used without issue, but the subject of interest will only be tracked
for a subset of the video. The video can also be reanalyzed with a
different algorithm, which normally show different idiosyncrasies.

In Figure 1, \texttt{TrackingBboxMethodLookup},
\texttt{TrackingBboxMethod}, \texttt{TrackingBbox},
\texttt{PersonBboxValid}, and \texttt{PersonBbox} are pipeline classes
(and tables) used to track and annotate the subject of interest.
\texttt{TrackingBboxMethodLookup} is a simple lookup table that
enumerates the implemented algorithms. \texttt{TrackingBboxMethod} are
manually entered rows that specify which algorithms to compute on which
videos. \texttt{TrackingBbox} is a computed class that calls the wrapper
for the selected algorithm indicated by the \texttt{TrackingBboxMethod}
entries with the specific video. We use this design pattern throughout
PosePipe and it is shown in Lst~\ref{lst:tracking_bbox}. Wrappers must
produce the bounding boxes tracklets in a standardized format, which
often takes minimal manipulation from output of the released
implementations. It is a list with entries for each frame and each entry
is itself a list of dictionaries. Each element in the dictionary
contains the tracklet ID, the bounding box coordinates and dimensions,
and the confidence the algorithm assigned to identifying the person in
that frame.

PosePipe provides wrappers for several algorithms. This includes
MMTrack\textsuperscript{\protect\hyperlink{ref-mmtracking_contributors_mmtracking_2020}{32}},
which provides a consistent API to several trackers and is under active
development. It also includes the released implementations of
DeepSort\textsuperscript{\protect\hyperlink{ref-wojke_simple_2017}{33}},
FairMOT\textsuperscript{\protect\hyperlink{ref-zhang_fairmot_2020}{34}},
TraDeS\textsuperscript{\protect\hyperlink{ref-wu_track_2021}{35}},
TransTrack\textsuperscript{\protect\hyperlink{ref-Sun2020}{36}}. We
implemented several tracking algorithms into our pipeline because the
optimal tracking algorithm for rehabilitation subjects is an open
question. Some algorithms seem to generalize poorly to rehabilitation
subjects and poor tracking precludes any subsequent analysis.

\begin{codelisting}

\caption{TrackingBbox Listing. This shows a standard DataJoint design pattern used throughout PosePipe to allow selecting specific algorithm implementations.}

\hypertarget{lst:tracking_bbox}{%
\label{lst:tracking_bbox}}%
\begin{Shaded}
\begin{Highlighting}[]

\KeywordTok{class}\NormalTok{ TrackingBbox(dj.Computed):}
\NormalTok{    definition }\OperatorTok{=} \StringTok{\textquotesingle{}\textquotesingle{}\textquotesingle{}}
\StringTok{    {-}\textgreater{} TrackingBboxMethod}
\StringTok{    {-}{-}{-}}
\StringTok{    tracks            : longblob}
\StringTok{    num\_tracks        : int}
\StringTok{    \textquotesingle{}\textquotesingle{}\textquotesingle{}}

    \KeywordTok{def}\NormalTok{ make(}\VariableTok{self}\NormalTok{, key):}

\NormalTok{        video }\OperatorTok{=}\NormalTok{ Video.get\_robust\_reader(key, return\_cap}\OperatorTok{=}\VariableTok{False}\NormalTok{)}

        \ControlFlowTok{if}\NormalTok{ (TrackingBboxMethodLookup }\OperatorTok{\&}\NormalTok{ key).fetch1(}\StringTok{\textquotesingle{}tracking\_method\_name\textquotesingle{}}\NormalTok{) }\KeywordTok{in} \StringTok{\textquotesingle{}MMTrack\_tracktor\textquotesingle{}}\NormalTok{:}
            \ImportTok{from}\NormalTok{ pose\_pipeline.wrappers.mmtrack }\ImportTok{import}\NormalTok{ mmtrack\_bounding\_boxes}
\NormalTok{            tracks }\OperatorTok{=}\NormalTok{ mmtrack\_bounding\_boxes(video, }\StringTok{\textquotesingle{}tracktor\textquotesingle{}}\NormalTok{)}
\NormalTok{            key[}\StringTok{\textquotesingle{}tracks\textquotesingle{}}\NormalTok{] }\OperatorTok{=}\NormalTok{ tracks}

        \ControlFlowTok{elif}\NormalTok{ (TrackingBboxMethodLookup }\OperatorTok{\&}\NormalTok{ key).fetch1(}\StringTok{\textquotesingle{}tracking\_method\_name\textquotesingle{}}\NormalTok{) }\OperatorTok{==} \StringTok{\textquotesingle{}MMTrack\_deepsort\textquotesingle{}}\NormalTok{:}
            \ImportTok{from}\NormalTok{ pose\_pipeline.wrappers.mmtrack }\ImportTok{import}\NormalTok{ mmtrack\_bounding\_boxes}
\NormalTok{            tracks }\OperatorTok{=}\NormalTok{ mmtrack\_bounding\_boxes(video, }\StringTok{\textquotesingle{}deepsort\textquotesingle{}}\NormalTok{)}
\NormalTok{            key[}\StringTok{\textquotesingle{}tracks\textquotesingle{}}\NormalTok{] }\OperatorTok{=}\NormalTok{ tracks}

        \ControlFlowTok{elif}\NormalTok{ (TrackingBboxMethodLookup }\OperatorTok{\&}\NormalTok{ key).fetch1(}\StringTok{\textquotesingle{}tracking\_method\_name\textquotesingle{}}\NormalTok{) }\OperatorTok{==} \StringTok{\textquotesingle{}MMTrack\_bytetrack\textquotesingle{}}\NormalTok{:}
            \ImportTok{from}\NormalTok{ pose\_pipeline.wrappers.mmtrack }\ImportTok{import}\NormalTok{ mmtrack\_bounding\_boxes}
\NormalTok{            tracks }\OperatorTok{=}\NormalTok{ mmtrack\_bounding\_boxes(video, }\StringTok{\textquotesingle{}bytetrack\textquotesingle{}}\NormalTok{)}
\NormalTok{            key[}\StringTok{\textquotesingle{}tracks\textquotesingle{}}\NormalTok{] }\OperatorTok{=}\NormalTok{ tracks}
        \ControlFlowTok{else}\NormalTok{:}
\NormalTok{            os.remove(video)}
            \ControlFlowTok{raise} \PreprocessorTok{Exception}\NormalTok{(}\SpecialStringTok{f"Unsupported tracking method: }\SpecialCharTok{\{}\NormalTok{key[}\StringTok{\textquotesingle{}tracking\_method\textquotesingle{}}\NormalTok{]}\SpecialCharTok{\}}\SpecialStringTok{"}\NormalTok{)}

\NormalTok{        track\_ids }\OperatorTok{=}\NormalTok{ np.unique([t[}\StringTok{\textquotesingle{}track\_id\textquotesingle{}}\NormalTok{] }\ControlFlowTok{for}\NormalTok{ track }\KeywordTok{in}\NormalTok{ tracks }\ControlFlowTok{for}\NormalTok{ t }\KeywordTok{in}\NormalTok{ track])}
\NormalTok{        key[}\StringTok{\textquotesingle{}num\_tracks\textquotesingle{}}\NormalTok{] }\OperatorTok{=} \BuiltInTok{len}\NormalTok{(track\_ids)}

        \VariableTok{self}\NormalTok{.insert1(key)}

        \CommentTok{\# remove the downloaded video to avoid clutter}
        \ControlFlowTok{if}\NormalTok{ os.path.exists(video):}
\NormalTok{            os.remove(video)}
\end{Highlighting}
\end{Shaded}

\end{codelisting}

\begin{figure}
\hypertarget{fig:annotation_gui}{%
\centering
\includegraphics[width=1.5625in,height=\textheight]{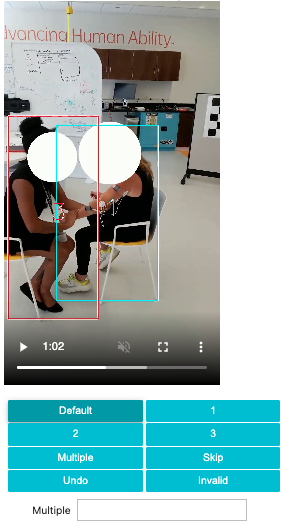}
\caption{Screenshot from GUI for annotation. Bounding boxes of tracklets
are shown with their identities, allowing the experimenter to
efficiently annotate the tracklets corresponding to the subject of
interest.}\label{fig:annotation_gui}
}
\end{figure}

After computing the tracklets, the next step is manual annotation of the
subject of interest. If there is only a single tracklet for the whole
video, it can be automatically selected. However, when there are
multiple tracklets, experimenter input is needed to properly identify
the one containing the subject of interest. We implemented a simple GUI
Fig.~\ref{fig:annotation_gui} that runs in a Jupyter Notebook and shows
the video with the tracklets overlaid (discussed more below). This
allows the user to select one of multiple tracklets that reflect the
subject of interest. This information is stored as rows in
\texttt{PersonBboxValid} with a subject ID. In most cases, we use a
subject ID of \(0\) to indicate the primary subject of interest or a
subject ID of -1 if the video was invalid for subsequent analysis (i.e.,
identity swaps or missed detection). However, this field could also
reflect unique subject IDs if using PosePipe alone to study multiple
individuals without additional experiment-specific DataJoint schemas, as
discussed more below. By having multiple \texttt{PersonBboxValid}
entries per video with different subject IDs, it is possible to track
multiple subjects in a video, such as for analyzing interaction between
patients and therapists. Based on this information, the
\texttt{PersonBbox} rows is populated and contains the validated
tracklets for individuals used for subsequent top-down algorithms.

\hypertarget{d-keypoint-detection}{%
\paragraph{2D Keypoint Detection}\label{d-keypoint-detection}}

Computing 2D keypoints using top-down algorithms depends on the
\texttt{PersonBbox} computed in the prior section and follows a similar
design pattern Fig.~\ref{fig:schema}. The supported list of 2D keypoint
algorithms are enumerated in the \texttt{TopDownMethodLookup} table and
the user selects specific algorithms to run by manually inserting the
corresponding rows into the \texttt{TopDownMethod} table. Populating
\texttt{TopDownPerson} runs a similar function as
Lst~\ref{lst:tracking_bbox}, where the selected algorithm determines
which wrapper is called. In each case, the wrappers use the video and
bounding box for each frame to extract a cropped portion of each image
with the subject of interest centered that will be passed to top down
algorithms. The output from the wrappers are a 3D array of dimension
frames \(\times\) num joints \(\times\) 3, where the last dimension has
the keypoint x and y coordinates and confidence estimates for each
keypoint. For any frame where there is no bounding box detected for the
subject of interest, the returned elements are \(\mathtt{NaN}\). For 2D
keypoint detection, we used the MMPose
Toolbox\textsuperscript{\protect\hyperlink{ref-mmpose_contributors_openmmlab_2020}{37}},
which provides a wide range of state-of-the-art neural network
architectures pretrained on multiple datasets.

\hypertarget{d-keypoint-lifting}{%
\paragraph{3D Keypoint Lifting}\label{d-keypoint-lifting}}

The sequence of 2D keypoints can be `lifted' to 3D with a number of
algorithms that have been trained on datasets of paired 2D-3D data in
order to resolve the ambiguity of the 2D observations and to constrain
the results to plausible 3-dimensional body configurations. Again, the
same design pattern is used with \texttt{LiftingMethodLookup} containing
the list of supported lifting methods and manually inserted rows in
\texttt{LiftingMethod} indicating which files and methods the user would
like to fill in when populating \texttt{LiftingPerson}. Wrapper
functions take the 2D keypoint array output from the previous step and
transform this into a 3D keypoint sequence with dimension frames
\(\times\) num joints \(\times\) 3, where the last dimensions are the
\(x\), \(y\) and \(z\) coordinates (lifting methods typically do not
produce a confidence estimate for each joint). We implemented a wrapper
for
GAST-Net\textsuperscript{\protect\hyperlink{ref-liu_graph_2020}{23}},
which lifts a sliding windows of 27 frames into 3D joint locations.

\hypertarget{smpl-fitting}{%
\paragraph{SMPL Fitting}\label{smpl-fitting}}

The SMPL body
model\textsuperscript{\protect\hyperlink{ref-Loper2015}{19}} is
parameterized with 10 parameters to describe the body shape and the
rotation at 23 joints, each of which has 3 degrees of freedom. There are
an additional 6 degrees of freedom to capture the overall body rotation
and position. It is worth noting that this is over-parameterized
compared to the typical human body. For example, a real knee and elbow
do not have 3 degrees of freedom. The SMPL-X
model\textsuperscript{\protect\hyperlink{ref-pavlakos_expressive_2019}{25}}
is an extension of SMPL with more degrees of freedom in the hands and
face to more expressively capture human movement.

PosePipe implements this class of algorithms with the same design
pattern used above: the \texttt{SMPLMethodLookup} table lists the
specific algorithms supported to estimate the parameters,
\texttt{SMPLMethod} is manually inserted into the database to indicate
that the user wants to analyze a particular video with a particular
algorithm, and \texttt{SMPLPerson} is populated using the bounding box
information and video to estimate the parameters using the selected
algorithms. Wrappers for various algorithms that estimate SMPL or SMPL-X
parameters are implemented, which use a similar approach to the 2D
keypoints by passing the cropped frames from the video to the selected
algorithms. The returned values from the wrappers include the model
type, body shape as a function of time, body pose as a function of time,
the 3D keypoint positions and 2D keypoints after reprojecting the 3D
keypoints into the image via the camera model, and the camera model used
during video acquisition. Different algorithms use inconsistent rotation
representation (e.g., rotation vectors passed through the Rodriguez
equation, quaternion rotations, rotation matrices, and 6D
representations), which are standardized by the wrappers to a rotation
vector. Additionally, the mathematical representation of the camera
model varies between algorithms from a weak perspective model (e.g.,
HMR\textsuperscript{\protect\hyperlink{ref-kanazawa_end_2018}{20}}) to a
full camera model
(e.g.,\textsuperscript{\protect\hyperlink{ref-kocabas_spec_2021}{38}}),
which is relevant because this can change the accuracy of the ultimate
inference and when producing the visualizations described below.

We implemented wrappers for several recent state-of-the-art SMPL or
SMPL-X parameter inference algorithms.
VIBE\textsuperscript{\protect\hyperlink{ref-kocabas_vibe_2020}{22}}
processes video sequences rather than individual frames to estimate SMPL
trajectories. PARE uses an attentional mechanisms to improve alignment
between body segments in the image and the body model reconstruction,
although processes frames
independently\textsuperscript{\protect\hyperlink{ref-kocabas_pare_2021}{39}}.
Expose\textsuperscript{\protect\hyperlink{ref-vedaldi_monocular_2020}{40}}
and a recent successor
PIXIE\textsuperscript{\protect\hyperlink{ref-feng_collaborative_2021}{41}}
processes individual frames to estimate parameters of an SMPL-X model,
which enables tracking finer scale changes like finger movements and
facial expression. PIXIE uses a similar attentional mechanism with an
SMPL-X model to improve the accuracy of hand tracking over
Expose\textsuperscript{\protect\hyperlink{ref-feng_collaborative_2021}{41}}.
ProHMR\textsuperscript{\protect\hyperlink{ref-kolotouros_probabilistic_2021}{42}}
produces a probability distribution over body poses for each frame which
can be fine-tuned based on the alignment to 2D keypoint estimates,
producing accurate fits at the expense of computation time. Finally,
HuMoR takes an alternative approach by optimizing the body model
parameter trajectory over time to accurately match the detected 2D
keypoints\textsuperscript{\protect\hyperlink{ref-rempe_humor_2021}{43}}.
We implemented this many algorithms because the field is advancing
incredibly
rapidly\textsuperscript{\protect\hyperlink{ref-tian_recovering_2022}{11}}
and different approach have different strengths and might be appropriate
depending on the question.

\hypertarget{visualization}{%
\subsubsection{Visualization}\label{visualization}}

Visualizing the output of different stages in the pipeline is critical
to identify failure modes and ensure these algorithms generalize
reliably to clinical populations. Because PosePipe is also designed for
data collected in clinical settings, a default behavior to preserve
privacy -- such as obscuring faces -- is an important feature.
OpenPose\textsuperscript{\protect\hyperlink{ref-Cao2016}{16}} is a
popular and efficient bottom-up HPE algorithm that detects keypoints,
including facial ones, for all people in a frame. Populating the
\texttt{OpenPose} table computes these keypoints for all videos and
allows populating the \texttt{BlurredVideo} table, which places a circle
over all faces detected by OpenPose. Like the \texttt{Video} table, the
video files from all visualizations are stored in external storage. The
\texttt{BlurredVideo} is then used to create overlays from other
algorithms such as computing \texttt{TrackingBboxVideo}, which show all
of the tracklets and are used for manual annotation
Fig.~\ref{fig:annotation_gui}, computing \texttt{TopDownPersonVideo} to
show the bounding box for the person of interest along with the detected
2D keypoints, or \texttt{SMPLPersonVideo} to show the estimated body
mesh shape overlaid on the video. The visualizations are implemented
with a consistent API that only needs to be provided a callback that
takes in the individual frame and frame index and then overlays the
desired information. This tool is also useful when producing
visualizations with additional analyses to estimate clinically relevant
parameters, for example gait event timing. It is also worth noting in
the PosePipe data schema Fig.~\ref{fig:schema} that the HPE algorithms
results do not depend on the visualizations, so they can be not computed
or deleted if space is a concern.

\hypertarget{organization-of-algorithms-and-weights}{%
\subsubsection{Organization of algorithms and
weights}\label{organization-of-algorithms-and-weights}}

There are several additional challenges to using multiple released
algorithms. Firstly, the code is often architected primarily for
evaluating performance on one or several datasets, and not for being
called from other software like a typical library. This includes the
challenge that multiple implementations may use identical directory
names, which can prevent Python from correctly identifying them if all
are added to the same path. Many algorithms also have a large number of
parameters that must be configured and passed into the script from the
command line, which further hinders calling them from external tools.
Secondly, deep learning algorithms are only partially specified by their
code and also require the weights determined after training algorithms.
These are often not distributed directly with the source code and must
be downloaded separately. In general, both the code and weights cannot
be universally distributed as a monolithic bundle as they may have
license requirements that must be respected. For example, one must
register and agree to the license to download the weights for the
SMPL/SMPL-X body models.

PosePipe reduces these challenges through two approaches. First, the
user configures PosePipe with a list of paths where each of the
algorithms has been downloaded locally. PosePipe then transiently adds
the specific algorithm to the path when running it and removes it when
complete. This prevents any namespace collision that would occur if all
algorithms were added to the path at once. To handle the weights, the
user must download them to a subdirectory in the PosePipe installation
(named 3rdparty). The wrappers specify the path within the subdirectory
and pass them to the initialized models, which avoids having to follow
algorithm-specific installation instructions for weights. Finally, the
wrappers also provide the list of configuration options compatible with
those weights and avoid the need for passing command line parameters to
external scripts.

\hypertarget{running-the-pipeline}{%
\subsubsection{Running the pipeline}\label{running-the-pipeline}}

The components described make up stages in a particular pipeline for
analyzing videos for HPE. An example of a common pipeline might look
like: 1) import videos, 2) run bounding box detection, 3) annotate
subject of interest, 4) run 2D keypoints detection, 5) run 3D keypoint
detection, 6) produce visualization. A specific pipeline can be
implemented using PosePipe with a short script, as shown in
Lst~\ref{lst:pipeline_example}. We refer readers to the DataJoint
documentation for details about the syntax. In this example, videos are
imported directly from a source directory. In most of our use of
PosePipe, videos to be analyzed have additional information pertaining
to the experimental question and we use additional DataJoint schemas to
organize the videos as part of the video import step. This script can
easily be modified to run in parallel on multiple GPUs by simply adding
\texttt{reserve=True} to the parameter of each populate method.

\begin{codelisting}

\caption{Example pipeline that produces an example from each of the algorithm classes.}

\hypertarget{lst:pipeline_example}{%
\label{lst:pipeline_example}}%
\begin{Shaded}
\begin{Highlighting}[]
\CommentTok{\# insert videos into database}
\NormalTok{video\_path }\OperatorTok{=} \StringTok{\textquotesingle{}/path/to/files\textquotesingle{}}
\NormalTok{videos\_files }\OperatorTok{=}\NormalTok{ os.listdir(video\_path)}
\ControlFlowTok{for}\NormalTok{ v }\KeywordTok{in}\NormalTok{ video\_files:}
\NormalTok{    insert\_local\_video(v, os.path.join(video\_path, v), video\_project}\OperatorTok{=}\StringTok{\textquotesingle{}PROJECT\_NAME\textquotesingle{}}\NormalTok{)}

\CommentTok{\# run tracking algorithms}
\NormalTok{keys }\OperatorTok{=}\NormalTok{ (Video }\OperatorTok{{-}}\NormalTok{ TrackingBboxMethod).fetch(}\StringTok{\textquotesingle{}KEY\textquotesingle{}}\NormalTok{)  }\CommentTok{\# find videos without bounding boxes computed}
\NormalTok{tracking\_method }\OperatorTok{=}\NormalTok{ (TrackingBboxMethodLookup }\OperatorTok{\&} \StringTok{\textquotesingle{}tracking\_method\_name="MMTrack"\textquotesingle{}}\NormalTok{) }\OperatorTok{\textbackslash{}}
\NormalTok{                  .fetch1(}\StringTok{\textquotesingle{}tracking\_method\textquotesingle{}}\NormalTok{)}
\NormalTok{TrackBboxMethod.insert([k.update(\{}\StringTok{\textquotesingle{}tracking\_method\textquotesingle{}}\NormalTok{: tracking\_method\}) }\ControlFlowTok{for}\NormalTok{ k }\KeywordTok{in}\NormalTok{ keys])}
\NormalTok{TrackingBbox.populate()  }\CommentTok{\# compute all of the tracking boxes}

\CommentTok{\# prepare blurred video for overlays}
\NormalTok{OpenPose.populate()}
\NormalTok{BlurredVideo.populate()}

\CommentTok{\# annotate videos}
\NormalTok{TrackingBboxVideo.populate()}

\CommentTok{\#\#\#\# run annotation GUI here, or automatically compute if only one person is in videos }\AlertTok{\#\#\#}

\NormalTok{PersonBbox.populate()  }\CommentTok{\# and compute the final bounding box for subjects of interest}

\CommentTok{\# compute 2D keypoints on videos}
\NormalTok{keys }\OperatorTok{=}\NormalTok{ (PersonBbox }\OperatorTok{{-}}\NormalTok{ TopDownMethod).fetch(}\StringTok{\textquotesingle{}KEY\textquotesingle{}}\NormalTok{) }\CommentTok{\# find videos without top down method selected}
\NormalTok{top\_down\_method }\OperatorTok{=}\NormalTok{ (TopDownMethodLookup }\OperatorTok{\&} \StringTok{\textquotesingle{}top\_down\_method\_name="MMPose"\textquotesingle{}}\NormalTok{).fetch1(}\StringTok{\textquotesingle{}top\_down\_method\textquotesingle{}}\NormalTok{)}
\NormalTok{TopDownMethod.insert([k.update(\{}\StringTok{\textquotesingle{}top\_down\_method\textquotesingle{}}\NormalTok{: top\_down\_method\}) }\ControlFlowTok{for}\NormalTok{ k }\KeywordTok{in}\NormalTok{ keys])}
\NormalTok{TopDownPerson.populate() }\CommentTok{\# analyze videos using selected algorithm}

\CommentTok{\# find videos waiting to be processed with lifting algorithm and run them}
\NormalTok{keys }\OperatorTok{=}\NormalTok{ (TopDownPerson }\OperatorTok{{-}}\NormalTok{ LiftingPersonMethod).fetch(}\StringTok{\textquotesingle{}KEY\textquotesingle{}}\NormalTok{)}
\NormalTok{lifting\_method }\OperatorTok{=}\NormalTok{ (LiftingMethodLookup }\OperatorTok{\&} \StringTok{\textquotesingle{}lifting\_method\_name="GastNet"\textquotesingle{}}\NormalTok{).fetch1(}\StringTok{\textquotesingle{}lifting\_method\textquotesingle{}}\NormalTok{)}
\NormalTok{LiftingPersonMethod.insert([k.update(\{}\StringTok{\textquotesingle{}lifting\_method\textquotesingle{}}\NormalTok{: lifting\_method\}) }\ControlFlowTok{for}\NormalTok{ k }\KeywordTok{in}\NormalTok{ keys])}
\NormalTok{LiftingPerson.populate() }\CommentTok{\# analyze videos using selected algorithm}

\CommentTok{\# find videos waiting to be processed with SMPL algorithm}
\NormalTok{keys }\OperatorTok{=}\NormalTok{ (PersonBbox }\OperatorTok{{-}}\NormalTok{ TopDownMethod).fetch(}\StringTok{\textquotesingle{}KEY\textquotesingle{}}\NormalTok{) }\CommentTok{\# find videos with no SMPL method selected}
\NormalTok{smpl\_method }\OperatorTok{=}\NormalTok{ (TopDownMethodLookup }\OperatorTok{\&} \StringTok{\textquotesingle{}top\_down\_method\_name="VIBE"\textquotesingle{}}\NormalTok{).fetch1(}\StringTok{\textquotesingle{}top\_down\_method\textquotesingle{}}\NormalTok{)}
\NormalTok{SMPLPersonMethod.insert([k.update(\{}\StringTok{\textquotesingle{}smpl\_method\textquotesingle{}}\NormalTok{: smpl\_method\}) }\ControlFlowTok{for}\NormalTok{ k }\KeywordTok{in}\NormalTok{ keys])}
\NormalTok{SMPLPerson.populate() }\CommentTok{\# analyze videos using selected algorithm}

\CommentTok{\# produce visualizations}
\NormalTok{TopDownPersonVideo.populate()}
\NormalTok{LiftingPersonVideo.populate()}
\NormalTok{SMPLPeronVideo.populate()}
\end{Highlighting}
\end{Shaded}

\end{codelisting}

\hypertarget{results}{%
\section{Results}\label{results}}

In our work, we found PosePipe is an effective tool for analyzing tens
of thousands of videos acquired in clinical settings. It greatly reduces
the barriers to testing emerging state-of-the-art algorithms for use in
HPE, as this can be done by only writing the wrapper to the novel
algorithms rather than creating \emph{de novo} pipelines between all the
requisite components. By storing the final outputs in DataJoint with an
associated experiment-specific schema, PosePipe makes subsequent
analysis for clinical variables of interest significantly easier.
Because this manuscript is primarily to introduce and describe PosePipe,
we will show some sample outputs, including errors, and note some
qualitative trends. However, we do not attempt a quantitative analysis
comparing specific algorithms.

\hypertarget{subject-tracking}{%
\paragraph{Subject tracking}\label{subject-tracking}}

\begin{figure}
\hypertarget{fig:bbox_example}{%
\centering
\includegraphics[width=4.16667in,height=\textheight]{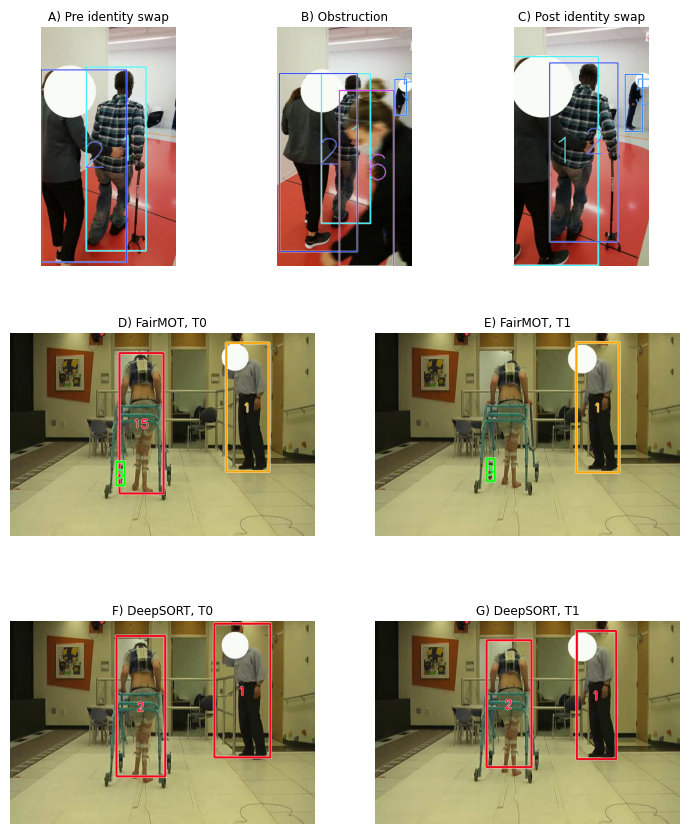}
\caption{Samples frames analyzed with different tracking algorithms.
A-C) Show an identity swap occurring after a brief obstruction. D and E)
are processed with FairMOT and shows how in some frames where the
subject of interest is not detected and false positives from the walker
wheels. F and G) analyze the same video without showing those
errors.}\label{fig:bbox_example}
}
\end{figure}

\begin{figure}
\hypertarget{fig:keypoint_examples}{%
\centering
\includegraphics[width=4.16667in,height=\textheight]{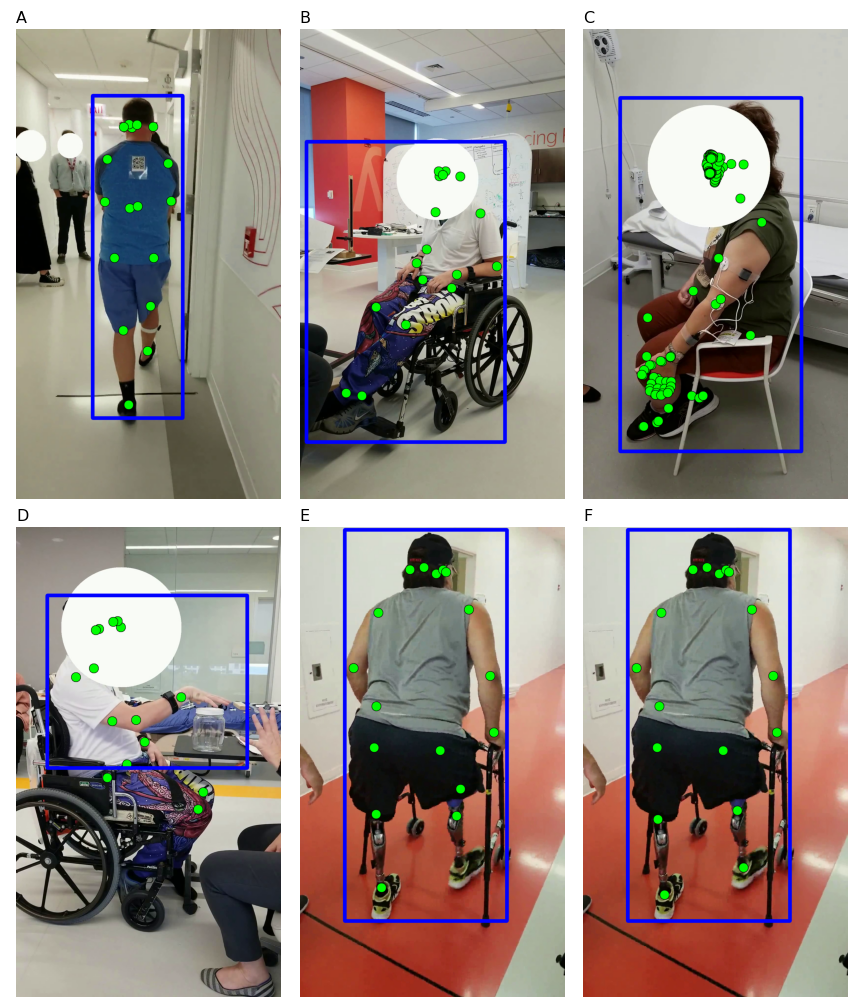}
\caption{A-C) Examples of top-down keypoint detection including several
examples of failures (D and E). C) also demonstrates WholeBody keypoints
including dense annotation of the hand and face. D) shows an example
where legs are missed due to bounding box detection missing the legs. E)
shows an example where the right prosthetic is missed. F) a DeepLabCut
model trained on prosthetic joints can resolve
this.}\label{fig:keypoint_examples}
}
\end{figure}

In most videos, especially videos with an unobstructed view of the whole
person throughout, we found that bounding box estimation produced a
single tracklet that uniquely mapped to the subject of interest.
However, several types of error were also noted
Fig.~\ref{fig:bbox_example}. The most problematic type of error was when
the tracklet swaps from tracking the person of interest to another
person. This is particularly common when a subject and therapist are
working closely together, and one briefly occludes the other. In these
cases, unless it is a brief tracklet that can simply be discarded, we
typically mark the tracking output as invalid and process the video with
a different algorithm. As different algorithms have different
idiosyncrasies this usually produces a usable output. For videos where
the subject is briefly occluded, multiple tracklets are sometimes
produced, but the annotation GUI allows stitching these together. In
rare cases, people would fail to be detected despite being clearly
visible. This seems to occur most commonly with assistive devices,
especially children with assistive devices, and was more pronounced when
using
FairMOT\textsuperscript{\protect\hyperlink{ref-zhang_fairmot_2020}{34}}.
Related to this, when people are sitting in wheelchairs, they are less
likely to be detected or the bounding box may only include the upper
torso (example shown in Fig.~\ref{fig:keypoint_examples}). From these
error observations, our typical initial tracking algorithm is
DeepSort\textsuperscript{\protect\hyperlink{ref-wojke_simple_2017}{33}}
as it quite reliably detects people and tracks them smoothly.

\hypertarget{top-down-2d-keypoint-detection}{%
\paragraph{Top Down 2D Keypoint
Detection}\label{top-down-2d-keypoint-detection}}

For top down 2D keypoint detection, we use two pretrained algorithms
from the MMPose
Toolbox\textsuperscript{\protect\hyperlink{ref-mmpose_contributors_openmmlab_2020}{37}}.
One algorithm detects the main body joints and several facial keypoints,
which uses an HRNet
architecture\textsuperscript{\protect\hyperlink{ref-Sun2019}{44}} and a
distribution-aware coordinate
representation\textsuperscript{\protect\hyperlink{ref-zhang_distribution-aware_2020}{45}}
trained on the COCO keypoints (ankles, knees, hips, shoulders, elbows,
wrists, eyes, ears and
nose)\textsuperscript{\protect\hyperlink{ref-lin_microsoft_2014}{46}}.
The other uses the same architecture but is trained on the
COCO-WholeBody\textsuperscript{\protect\hyperlink{ref-jin_whole-body_2020}{47}}
dataset, which includes 68 facial keypoints and 42 on the hands to
produce much more fine-grained tracking. We noted that people
interacting closely also can introduce errors at this step. These
algorithms are fairly reliable, provided the bounding box was detected
accurately and the joint is not visually obscured. However, errors do
occur when limbs appear different than able bodied adults, such as thin
limbs wearing braces or some prosthetic users. In these cases, a custom
model can be trained with
DeepLabCut\textsuperscript{\protect\hyperlink{ref-Mathis2018}{48},\protect\hyperlink{ref-Nath2019}{49}},
which allows manually annotating the location of prosthetic joints. A
custom algorithm can be created and entered in
\texttt{TopDownPersonLookup} and can then be used to combine the
keypoints from MMPose of the intact joints with the prosthetic joint
locations from DeepLabCut. Examples of these algorithms and examples of
errors are shown in Fig.~\ref{fig:keypoint_examples}.

\hypertarget{d-keypoint-lifting.}{%
\paragraph{3D Keypoint Lifting.}\label{d-keypoint-lifting.}}

\begin{figure}
\hypertarget{fig:lifting_example}{%
\centering
\includegraphics[width=4.16667in,height=\textheight]{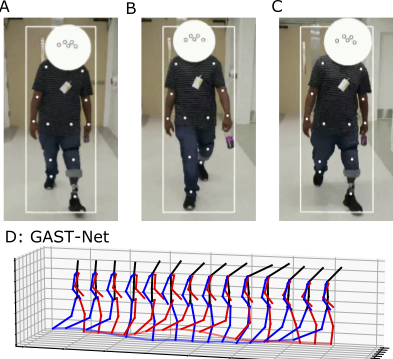}
\caption{A-C) Several frames of someone walking with bounding box and 2D
keypoints that correspond to the beginning, middle and end of the
trajectory shown below. D) The 3D skeleton trajectory lifted using
GAST-Net showing the gait cycle. Note that the skeleton appears
accurate. despite being viewed orthogonally to the camera perspective.
Red corresponds to the left side and blue the
right.}\label{fig:lifting_example}
}
\end{figure}

GAST-Net\textsuperscript{\protect\hyperlink{ref-liu_graph_2020}{23}}
produces realistic appearing 3D keypoint trajectories, provided the
bounding box and 2D keypoints were visible and detected accurately.
These trajectories are also quite smooth, likely due to the combination
of information over multiple frames. However, because of the scale
ambiguity from pure 2D keypoints, the joint locations are only relative
and do not scale to the individual. Example
Fig.~\ref{fig:lifting_example} shows an example stick figure
reconstruction of a subject walking, clearly showing the concordance
between the frames and body positions.

\hypertarget{smpl-fitting-1}{%
\paragraph{SMPL Fitting}\label{smpl-fitting-1}}

\begin{figure}
\hypertarget{fig:smpl_examples}{%
\centering
\includegraphics[width=4.16667in,height=\textheight]{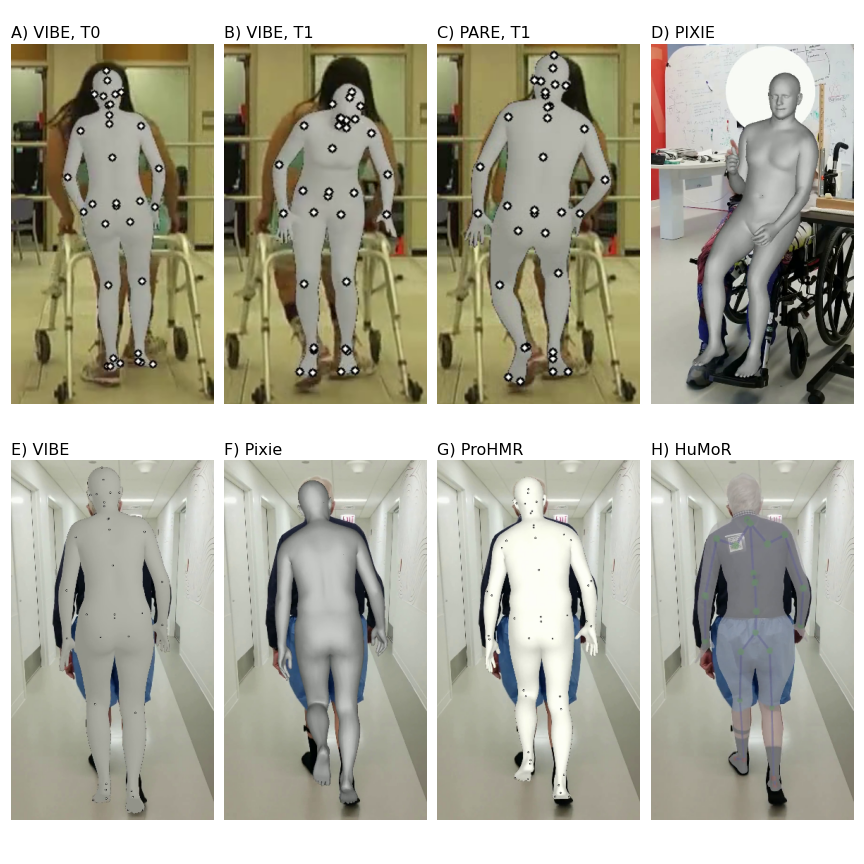}
\caption{Examples estimating SMPL/SMPL\_X meshes. A,B) Two frames
processed with VIBE showing confusion about the direction of the
subject. C) In comparison, PARE did not have this error and more
accurately aligned with the limbs. D) PIXIE can capture detail hand
gestures, such as a thumbs up. E,F) Both VIBE and PIXIE do not always
capture the leg placement properly, although the later still captures
fine hand movements. G) ProHMR can improve the alignment of the mesh
joints with the image. H) HuMoR shows particularly good tracking of the
joints with smooth trajectories.}\label{fig:smpl_examples}
}
\end{figure}

Estimating SMPL parameters produced mixed results. In many cases, the
results were promising, but sometimes contained notable examples of
brittleness (i.e.~sensitivity to irrelevant features)
Fig.~\ref{fig:smpl_examples}. For example, with
VIBE\textsuperscript{\protect\hyperlink{ref-kocabas_vibe_2020}{22}} we
noted instances where the presence of assistive device caused
significant errors.
PARE\textsuperscript{\protect\hyperlink{ref-kocabas_pare_2021}{39}}
makes estimation of body models more robust to occlusion through an
attentional mechanism, and seemed to reduce this sensitivity.
Expose\textsuperscript{\protect\hyperlink{ref-vedaldi_monocular_2020}{40}}
and
PIXIE\textsuperscript{\protect\hyperlink{ref-feng_collaborative_2021}{41}}
uses the SMPL-X model and estimates parameters for the shape of the
hand. PIXIE is more recent and produces more consistent results,
although both outputs contain high frequency jitter in the details and
particularly struggled when the video contained motion blur artifacts or
additional hands nearby. Another approach that improves the alignment of
the mesh output with limbs in the image is
ProHMR\textsuperscript{\protect\hyperlink{ref-kolotouros_probabilistic_2021}{42}}.
It infers a probability distribution over poses for each frame which can
be combined with the robustly estimated 2D keypoints to compute
optimized SMPL parameters that are consistent with both the image and
keypoints. This has the benefits of capturing nuance that is not
contained in the 2D keypoint location of major body joints (e.g., wrist
supination and pronation), while reducing gratuitous errors. However,
the additional optimization step takes several seconds per frame, making
analyzing many videos with this algorithm very time consuming. It also
analyzes individual frames independently, so can contain jitter.
HuMoR\textsuperscript{\protect\hyperlink{ref-rempe_humor_2021}{43}} is
another algorithm that uses an optimization approach to 2D keypoints and
specifically optimizes a trajectory over time. The outputs from HuMoR
are very smooth and well aligned to the body, although in rare cases it
fails to converge to a good solution. It also can take multiple hours to
optimize the pose trajectory for a video.

\hypertarget{experiment-specific-schema}{%
\paragraph{Experiment-specific
schema}\label{experiment-specific-schema}}

The pipeline can be easily used with experiment specific DataJoint
schemas that help perform and organize subsequent analyses
Fig.~\ref{fig:pd_schema}. In this example case shown, the
\texttt{Subject} table contains rows with information for each subject,
and the \texttt{Activity} table contains rows for each activity they
performed and is linked to entries in the \texttt{Video} table. SMPL
trajectories are computed using the pipeline we have described and
stored in \texttt{SMPLPerson} . The \texttt{FtnStatistics} and
\texttt{RamStatistics} classes compute summary statistics of specific
activities (in this case performing finger to nose movements or rapid
alternating movements). The link to the \texttt{Activity} table ensures
the specific analysis are only performed on appropriate activities and
makes it easy to retrieve the right data. For example to retrieve the
frequency of movement at different time points for an individual, it is
as simple as:

\begin{verbatim}
timepoints, frequency = (FtnStatistics & 'subject_id=5') \
                         .fetch('timepoints', 'frequency')
\end{verbatim}

\begin{figure}
\hypertarget{fig:pd_schema}{%
\centering
\includegraphics[width=2.08333in,height=\textheight]{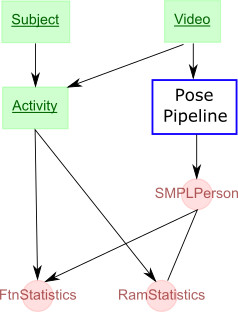}
\caption{Examples of an experiment specific schema. The complete
PosePipe diagram from Fig 1 is abstracted into the blue box and
additional nodes indicate the data organization for a particular
experiment}\label{fig:pd_schema}
}
\end{figure}

\hypertarget{discussion}{%
\section{Discussion}\label{discussion}}

We developed PosePipe as a simplified, easy to use method for HPE
analysis on large volumes of videos acquired in clinical situations. We
find this system makes it much easier to test and compare different
algorithms on videos within a consistent framework. This comes from
using DataJoint as the framework for
PosePipe\textsuperscript{\protect\hyperlink{ref-yatsenko_datajoint_2015}{29}},
which stores the results of each stage of the HPE pipeline in a database
while managing all of the computational dependencies and ensuring data
integrity. As shown by Lst~\ref{lst:pipeline_example}, videos can be
analyzed in a custom pipeline with only a short script. The tables in
DataJoint enforce a consistent data format to represent each class of
algorithm (e.g., bounding box tracklet computation) and newly released
algorithms can be supported simply by writing a wrapper to the algorithm
compatible with this data format.

The accuracy of the outputs from PosePipe depends upon the accuracy of
specific implementations and their performance when combined. Our goal
with this work is very intentionally \emph{not} to produce or train a
state-of-the-art algorithm, but to build a tool that advances the use of
new HPE algorithms for clinical and translation research, with a
particular focus on rehabilitation. A systematic evaluation of the
accuracy of specific components applied to clinical populations, while
extremely important, is outside the scope of this work. However, our
qualitative results offer several tips and warnings for using various
HPE algorithms for clinical and translation research. Our broadest and
most strongly recommended tip is the importance of rendering and
reviewing the visualizations of different algorithms, rather than
blithely trusting the outputs. While HPE tools are advancing rapidly,
they are still not completely reliable; this is particularly true when
they are applied to clinical populations that may be systematically
different from the data the algorithms were trained on.

\hypertarget{subject-tracking-1}{%
\paragraph{Subject tracking}\label{subject-tracking-1}}

For tracking algorithms, identity swaps of the tracklets are most common
when two individuals are in close proximity (e.g., patient and
therapist) and can render the output unusable in some cases. Frequently,
reprocessing a video with identity swaps with a different algorithm
produces a better result. Fragmented tracklets are a lesser problem, but
these increase the time needed to manually annotate the subject of
interest in videos and are an obstacle to high throughput, fully
automated approaches. Detection gaps commonly occur when an individual
is transiently not detected, either due to occlusion or because of an
algorithm error. These failures to detect seem to occur more frequently
when people use an assistive device and this was particularly notable
with
FairMOT\textsuperscript{\protect\hyperlink{ref-zhang_fairmot_2020}{34}}.
Further, the use of a wheelchair sometimes results in bounding boxes
that miss the subject's legs. Whether one algorithm is consistently the
most reliable for rehabilitation patients remains an open question. Our
current impression is that each algorithm has different idiosyncrasies.
However, we are optimistic that the continual advances in tracking for
HPE will continue to lessen this problem
(e.g.\textsuperscript{\protect\hyperlink{ref-rajasegaran_tracking_2021}{50},\protect\hyperlink{ref-yuan_glamr_2021}{51}}),
with the caveat that underrepresentation of people with disabilities in
the publicly available training datasets may bound the performance when
they are applied for those who might benefit the most from this
technology.

\hypertarget{top-down-2d-keypoints}{%
\paragraph{Top down 2D keypoints}\label{top-down-2d-keypoints}}

For 2D keypoint detection, we used the MMPose
toolbox\textsuperscript{\protect\hyperlink{ref-mmpose_contributors_openmmlab_2020}{37}},
which offers many benefits including a standardized API to use multiple
different cutting-edge architectures trained on a range of datasets and
the fact it is actively maintained, with new algorithms routinely added.
In our results, we have primarily utilized their implementation of a
HRNet trained on the COCO body keypoints (ankles, knees, hips,
shoulders, elbows, wrists, eyes, ears and
nose)\textsuperscript{\protect\hyperlink{ref-lin_microsoft_2014}{46}}
and find this generally works quite well for most cases tested. It
primarily fails when tested on people with limbs that do not resemble an
intact limb of an able-bodied person, such as amputees using a
prosthetic device and particularly for people with more proximal or
bilateral amputations. In this case, the visual difference between a
prosthetic and intact limb makes it unlikely that any contemporary
algorithm trained on able-bodied data alone will succeed in this task,
because the neural networks trained to solve this task are designed to
recognize relatively low level visual properties rather than reason
about the functional homology of a prosthetic and intact limb. As such,
we found it was occasionally necessary to use a tool like
DeepLabCut\textsuperscript{\protect\hyperlink{ref-Mathis2018}{48},\protect\hyperlink{ref-Nath2019}{49}},
which allows training a custom detection algorithm after manually
annotating prosthetic joints in some videos. We have previously found
this is also required when videos do not include the upper torso of
people\textsuperscript{\protect\hyperlink{ref-lonini_video-based_2022}{52}}.
In general, we see 2D keypoint trajectories an impoverished
representation for the true 3D biomechanical movement a subject performs
in the world. This is unfortunate as they are some of the most robust
HPE algorithms available. In many works, these estimates are referred to
as a 2D pose, but avoid this nomenclature as the true pose cannot be
recovered directly from these estimates.

\hypertarget{lifted-3d-joint-locations}{%
\paragraph{Lifted 3D joint locations}\label{lifted-3d-joint-locations}}

When the 2D keypoints were accurately detected and the person was fully
visible, the lifted 3D keypoint trajectories appeared accurate. It is
worth noting that lifting algorithms are not calibrated to match the
height of an individual, which is particularly relevant when analyzing
the movements of children who will be scaled towards an adult. As
mentioned, 3D joint locations cannot be recovered analytically from 2D
keypoints. This is because any 3D location along an epipolar line from
the camera would project to the same point in the image plane, creating
a fundamental ambiguity. Lifting methods address this ambiguity by not
just trying to find 3D points consistent with these epipolar lines, but
also those consistent with plausible human configurations. Essentially,
they learn a priori over poses from the training data. This raises the
concern that they may exhibit biases when tested on people who move
differently than the able-bodied population these algorithms were
trained on, such as people with range of motion restrictions or people
with dystonic cerebral palsy or other movement disorders.

\hypertarget{smpl-methods}{%
\paragraph{SMPL Methods}\label{smpl-methods}}

Algorithms that estimate parameters of body-models such as
SMPL\textsuperscript{\protect\hyperlink{ref-Loper2015}{19}} or
SMPL-X\textsuperscript{\protect\hyperlink{ref-pavlakos_expressive_2019}{25}}
are particularly promising as they provide an inference that is closer
to the biomechanical understanding than 3D joint locations by finding
the joint angles or pose that recreates the body configuration in the
image. However, the SMPL/SMPL-X models were developed with a focus on
realistic computer graphics and producing an accurate external body
shape rather than with a detailed biomechanical focus, so there is not a
1:1 mapping from these parameters to the International Standard of
Biomechanics recommended
descriptions\textsuperscript{\protect\hyperlink{ref-wu_isb_2002}{26},\protect\hyperlink{ref-Wu2005}{27}},
although we have previously shown these can be computed from the SMPL
parameters for the
arm\textsuperscript{\protect\hyperlink{ref-Cotton2020}{28}}.

A limitation of these algorithms is they sometimes have significant
errors, particularly in the presence of occlusions. Because of this, we
find visualizing the outputs prior to using them is a critical step.
More recent methods that include attentional mechanisms, like
PARE\textsuperscript{\protect\hyperlink{ref-kocabas_pare_2021}{39}}, are
more robust to occlusions. With the exception of
VIBE\textsuperscript{\protect\hyperlink{ref-kocabas_vibe_2020}{22}} and
HuMoR\textsuperscript{\protect\hyperlink{ref-rempe_humor_2021}{43}}, the
methods we tested analyze frames independently which also results in
increased jitter. Most also do not typically produce confidence
estimates to know when they are accurate and can be trusted. An
exception is
ProHMR\textsuperscript{\protect\hyperlink{ref-kolotouros_probabilistic_2021}{42}}
produces a probability distribution over poses that can be refined using
detected keypoints. This produces very promising results for HPE
analysis as it reduces these inconsistencies by optimizing the
parameters to align the mesh to the keypoints, but the time to run this
is quite limiting and it tends to have a fair amount of jitter between
frames.
Expose\textsuperscript{\protect\hyperlink{ref-vedaldi_monocular_2020}{40}}
and
PIXIE\textsuperscript{\protect\hyperlink{ref-feng_collaborative_2021}{41}}
are useful for questions involving hand function with the later being
more accurate in our experience, but these also tend to have some jitter
that limits analyzing detailed hand trajectories. If expressive hand
movement isn't required,
HuMoR\textsuperscript{\protect\hyperlink{ref-rempe_humor_2021}{43}}
produces excellent results that are very smooth, although the time
required to optimize the trajectories can be prohibitive.

\hypertarget{need-for-systematic-evaluation}{%
\paragraph{Need for systematic
evaluation}\label{need-for-systematic-evaluation}}

With the increasing use of deep learning systems in medicine, there is a
concern that they function as a black box and the need for Explainable
Artificial Intelligence (XAI) has been
emphasized\textsuperscript{\protect\hyperlink{ref-holzinger_causability_2019}{53},\protect\hyperlink{ref-amann_explainability_2020}{54}}.
XAI has multiple components and
perspectives\textsuperscript{\protect\hyperlink{ref-vilone_notions_2021}{55}},
with one notion being that intermediate steps produce outputs that can
be meaningfully interpreted and reviewed, as well as understanding the
influence of those intermediate steps on the final output. The
trajectory of a subject's movement -- regardless of the specific
representation -- is typically an input used to compute clinically
pertinent metrics (e.g., gait cadence and deviation
index\textsuperscript{\protect\hyperlink{ref-Kidzinski2020}{13}}, gait
temporal
parameters\textsuperscript{\protect\hyperlink{ref-lonini_video-based_2022}{52}},
or Parkinson's Disease motor symptom
severity\textsuperscript{\protect\hyperlink{ref-lu_quantifying_2021}{56}}).
They also have the benefit of being very interpretable and can be easily
checked with visualizations. Thus, to improve the explainability of
these systems, it is critical to know how accurate the pose estimate
inputs are and how that accuracy influences overall algorithm
reliability.

Given the potential benefit of these algorithms for rehabilitation
research and outcome measures, there is a substantial need for
large-scale validation studies of these algorithms when applied to
clinical populations. Quantitatively evaluating the accuracy of HPE
algorithms on any clinical populations is challenging because there are
few datasets that reflect real world use cases with ground truth
annotation. These are challenging to collect because ground truth using
optical motion tracking requires bringing subjects into a motion
analysis laboratory. Advances in wearable sensors or pose estimation
with multiple cameras could reduce this barrier while producing
sufficiently accurate annotation.

Additionally, these validation studies are important for understanding
how algorithmic
fairness\textsuperscript{\protect\hyperlink{ref-mehrabi_survey_2021}{57}--\protect\hyperlink{ref-Chouldechova2018}{59}}
interacts with people with
disabilities\textsuperscript{\protect\hyperlink{ref-trewin_considerations_2019}{60}}.
For example, our observations that the use of assistive devices, such as
a walker, seems to reduce the probability that a person (particularly a
child) is detected or that in some cases prosthetic limbs are poorly
tracked. One contribution to this type of bias may be the
underrepresentation of people with disabilities in the datasets used to
train the algorithms. It is also important to note that dataset bias is
only one source of algorithmic bias. Because applying these algorithms
to clinical populations may be the situation where they can provide the
most societal benefit, it is critical to understand and address sources
of any systematic errors.

\hypertarget{facilitating-analysis-of-hpe-outputs}{%
\subsection{Facilitating analysis of HPE
outputs}\label{facilitating-analysis-of-hpe-outputs}}

PosePipe makes analysis of the HPE outputs for the primary research
questions much easier. DataJoint provides an efficient mechanism to
allow multiple users to easily access the outputs from multiple
computers (if enabled) and benefits from all the development making
MySQL highly performant. DataJoint provides a pythonic API to retrieve
the specific data desired that maps to MySQL queries. For example,
retrieving a SMPL pose trajectory for a specific video over a remote
connection takes a negligible amount of time. This also avoids recurring
challenges associated with using a file system for data organization.
DataJoint allows granular user access controls, including assigning
users read-only access to certain tables, which can prevent them from
accidentally deleting raw data. Controlling access to the external data
stores also determines who can or cannot view the raw videos and/or
visualizations, which may help to restrict access to identifiable
information.

In many cases, videos are collected with additional metadata and
experimental information that is critical for subsequent analysis of the
HPE results from those videos. In these cases, we recommend designing an
experiment-specific DataJoint schema that organizes the videos
accordingly and is used to insert them into PosePipe (as opposed to
Lst~\ref{lst:pipeline_example} which imports them directly from a
directory organized only by filename). Experiment-specific schemas
provide a tremendous benefit. For example, there may be a table
containing demographic or clinical information about subjects and
another table that computes a clinically relevant feature from the 3D
keypoints, such as walking speed. In this case, a one-line query can
return the walking speed of a subject associated with their demographic
data and the data of measurements in a Pandas
dataframe\textsuperscript{\protect\hyperlink{ref-the_pandas_development_team_pandas-devpandas_2020}{61}}.
We defer further specific examples to subsequent manuscripts using
outputs from PosePipe.

\hypertarget{limitations}{%
\subsection{Limitations}\label{limitations}}

In addition to limitations on accuracy from the individual components
just discussed, PosePipe itself has several limitations. Setup requires
several steps. These include setting up a DataJoint database, which is
fairly straightforward using the provided Docker container. It also
includes downloading the code and model weights for each of the
algorithms the user wishes to apply to videos. Detailed installation
instructions for all steps are provided in the PosePipe repository.

The facial blurring is an important default privacy-preserving feature,
but is only as reliable as the face detection algorithm used (currently
OpenPose). This makes additional manual review an important step before
releasing any videos requiring anonymization.

The DataJoint model also introduces some friction when interacting with
experiment-specific schemas. For example, videos may be conceptually
organized as children of other tables, such as a table of experimental
sessions which itself is a descendent from a table of clinical subjects
with demographic information. However, when working with multiple
experiments it is not possible for some \texttt{Video} rows to have a
parent from one experimental table and in other cases a different parent
table. One future option is to adopt DataJoint
Elements\textsuperscript{\protect\hyperlink{ref-yatsenko_datajoint_2021}{62}},
which allows creating reusable pipelines that can be attached to
different experimental tables.

\hypertarget{future-directions}{%
\subsection{Future Directions}\label{future-directions}}

Consistent with our goal to facilitate HPE in research, an important
future step is to simplify the installation process. We envision doing
so by developing a Docker container with all the required files for a
minimal pipeline, including the DataJoint database and a minimal set of
algorithms that could be distributed consistent with their licenses.
This will likely heavily leverage implementations provided by
MMPose\textsuperscript{\protect\hyperlink{ref-mmpose_contributors_openmmlab_2020}{37}}
and
MMTrack\textsuperscript{\protect\hyperlink{ref-mmtracking_contributors_mmtracking_2020}{32}},
which provide installable libraries and numerous competitive pretrained
algorithms that can be accessed with a straightforward API.

Further, we hope to support new classes of algorithms that are becoming
available. We are particularly excited about approaches that integrate
physics-based biomechanical modeling to constrain estimates to more
physically plausible ones while also inferring joint torques and ground
reaction
forces\textsuperscript{\protect\hyperlink{ref-shimada_physcap_2020}{66}},
although these will also require substantial amounts of validation. We
also plan to include action recognition algorithms in future
revisions\textsuperscript{\protect\hyperlink{ref-gu_ava_2018}{67}}.

Additionally, we will use PosePipe to perform systematic evaluations on
the performance of specific algorithms on videos of rehabilitation
subjects, as described above. The lack of appropriate datasets with
ground truth annotation makes this task difficult, but does not preclude
having human raters evaluate the quality of outputs, which would still
allow quantitative comparison of different algorithms. We are currently
developing tools to enable collection of simultaneous video and wearable
sensor data and video data with multiple cameras, which can provide
additional signals for both evaluation and training of algorithms on
data acquired in the
clinic\textsuperscript{\protect\hyperlink{ref-Cotton2020}{28},\protect\hyperlink{ref-Cotton2019}{68}}.

Most importantly, we will use our PosePipe tool to further our ultimate
goal: to develop clinically useful tools and outcome measures that can
be used in wide ranging clinical contexts. High-quality HPE analysis,
even if it produces near-perfect biomechanical understanding, will need
additional analysis to produce clinically useful measures. As an
example, when performing gait analysis, the first step is detecting the
timing of gait events, such as foot contact and toe off, and then, after
aligning the gait cycle, additional statistics can be computed and
compared to normative
datasets\textsuperscript{\protect\hyperlink{ref-richard_whittles_2012}{69},\protect\hyperlink{ref-whittle_clinical_1996}{70}}.
Thus, advancing the clinical utility of HPE will also require developing
similar high-level clinical measures and validating the properties of
these measures on clinical populations.

\hypertarget{conclusion}{%
\section{Conclusion}\label{conclusion}}

We introduce PosePipe, an open-source tool based on DataJoint that makes
it easy to implement pipelines to analyze videos of human movement
acquired in clinical situations using cutting-edge algorithms for human
pose estimation. It supports several released implementations from
different classes of algorithms including bounding box tracklet
computation to track an individual in videos, top-down 2D keypoint
estimation, lifting 2D keypoints to 3D joint locations, and estimating
the parameters of SMPL/SMPL-X body models. We anticipate that this tool
will facilitate the use of these algorithms in clinical and
translational research for movement science and rehabilitation and help
enable much needed systematic evaluation of their performance when
tested on clinical populations.

\hypertarget{acknowledgment}{%
\paragraph{Acknowledgment}\label{acknowledgment}}

This work was generously supported by the Research Accelerator Program
of the Shirley Ryan AbilityLab. We would like to thank Tasos Karakostas,
Arun Jayaraman, and Anothony Cimorelli for data used in the development
of this pipeline and productive discussions. We would also like to thank
Dimitri Yatsenko, Meghan OConnell and Kyle Embry for constructive
feedback on the manuscript and code.

\hypertarget{author-contribution}{%
\paragraph{Author Contribution}\label{author-contribution}}

RJC developed PosePipe, analyzed the videos used as examples in this
work, and wrote the manuscript.

\hypertarget{references}{%
\section*{References}\label{references}}
\addcontentsline{toc}{section}{References}

\hypertarget{refs}{}
\begin{CSLReferences}{0}{0}
\leavevmode\vadjust pre{\hypertarget{ref-poitras_validity_2019}{}}%
\CSLLeftMargin{1. }
\CSLRightInline{Poitras, I. \emph{et al.}
\href{https://doi.org/10.3390/s19071555}{Validity and Reliability of
Wearable Sensors for Joint Angle Estimation: A Systematic Review}.
\emph{Sensors (Basel)} \textbf{19}, 1555 (2019).}

\leavevmode\vadjust pre{\hypertarget{ref-Cuesta-Vargas2010}{}}%
\CSLLeftMargin{2. }
\CSLRightInline{Cuesta-Vargas, A. I., GalÃ¡n-Mercant, A. \& Williams, J.
M. \href{https://doi.org/10.1179/1743288X11Y.0000000006}{The use of
inertial sensors system for human motion analysis}. \emph{Physical
Therapy Reviews} \textbf{15}, 462--473 (2010).}

\leavevmode\vadjust pre{\hypertarget{ref-Filippeschi2017}{}}%
\CSLLeftMargin{3. }
\CSLRightInline{Filippeschi, A. \emph{et al.}
\href{https://doi.org/10.3390/s17061257}{Survey of motion tracking
methods based on inertial sensors: A focus on upper limb human motion}.
\emph{Sensors (Basel)} \textbf{17}, (2017).}

\leavevmode\vadjust pre{\hypertarget{ref-Zheng2020}{}}%
\CSLLeftMargin{4. }
\CSLRightInline{Zheng, C. \emph{et al.} \emph{Deep Learning-Based Human
Pose Estimation: A Survey}. 663--676
\url{https://github.com/zczcwh/DL-HPE} (2020).}

\leavevmode\vadjust pre{\hypertarget{ref-Kwakkel2019a}{}}%
\CSLLeftMargin{5. }
\CSLRightInline{Kwakkel, G. \emph{et al.} Standardized measurement of
quality of upper limb movement after stroke: Consensus-based core
recommendations from the Second Stroke Recovery and Rehabilitation
Roundtable. \emph{International journal of stroke : official journal of
the International Stroke Society} 1747493019873519 (2019)
doi:\href{https://doi.org/10.1177/1747493019873519}{10.1177/1747493019873519}.}

\leavevmode\vadjust pre{\hypertarget{ref-Seethapathi2019}{}}%
\CSLLeftMargin{6. }
\CSLRightInline{Seethapathi, N., Wang, S., Saluja, R., Blohm, G. \&
Kording, K. P. \href{http://arxiv.org/abs/1907.10226}{Movement science
needs different pose tracking algorithms}. \emph{arXiv} (2019).}

\leavevmode\vadjust pre{\hypertarget{ref-Parks2019}{}}%
\CSLLeftMargin{7. }
\CSLRightInline{Parks, M. T., Wang, Z. \& Siu, K.-C.
\href{https://doi.org/10.1093/ptj/pzz097}{Current Low-Cost Video-Based
Motion Analysis Options for Clinical Rehabilitation: A Systematic
Review.} \emph{Physical therapy} \textbf{99}, 1405--1425 (2019).}

\leavevmode\vadjust pre{\hypertarget{ref-Needham2021}{}}%
\CSLLeftMargin{8. }
\CSLRightInline{Needham, L. \emph{et al.} \emph{Human Movement Science
in The Wild: Can Current Deep-Learning Based Pose Estimation Free Us
from The Lab?} 2021.04.22.440909
\url{https://www.biorxiv.org/content/10.1101/2021.04.22.440909v1} (2021)
doi:\href{https://doi.org/10.1101/2021.04.22.440909}{10.1101/2021.04.22.440909}.}

\leavevmode\vadjust pre{\hypertarget{ref-Liu2021_Pose}{}}%
\CSLLeftMargin{9. }
\CSLRightInline{Liu, W., Bao, Q., Sun, Y. \& Mei, T.
\href{http://arxiv.org/abs/2104.11536}{Recent Advances in Monocular 2D
and 3D Human Pose Estimation: A Deep Learning Perspective}.
\emph{arXiv:2104.11536 {[}cs{]}} (2021).}

\leavevmode\vadjust pre{\hypertarget{ref-chen_monocular_2020}{}}%
\CSLLeftMargin{10. }
\CSLRightInline{Chen, Y., Tian, Y. \& He, M.
\href{https://doi.org/10.1016/j.cviu.2019.102897}{Monocular human pose
estimation: A survey of deep learning-based methods}. \emph{Computer
Vision and Image Understanding} \textbf{192}, 102897 (2020).}

\leavevmode\vadjust pre{\hypertarget{ref-tian_recovering_2022}{}}%
\CSLLeftMargin{11. }
\CSLRightInline{Tian, Y., Zhang, H., Liu, Y. \& Wang, L.
\href{http://arxiv.org/abs/2203.01923}{Recovering 3D Human Mesh from
Monocular Images: A Survey}. \emph{arXiv:2203.01923 {[}cs{]}} (2022).}

\leavevmode\vadjust pre{\hypertarget{ref-sato_quantifying_2019}{}}%
\CSLLeftMargin{12. }
\CSLRightInline{Sato, K., Nagashima, Y., Mano, T., Iwata, A. \& Toda, T.
\href{https://doi.org/10.1371/journal.pone.0223549}{Quantifying normal
and parkinsonian gait features from home movies: Practical application
of a deep learning-based 2D pose estimator}. \emph{PLoS One}
\textbf{14}, e0223549 (2019).}

\leavevmode\vadjust pre{\hypertarget{ref-Kidzinski2020}{}}%
\CSLLeftMargin{13. }
\CSLRightInline{Kidzinski, L. \emph{et al.}
\href{https://doi.org/10.1038/s41467-020-17807-z}{Deep neural networks
enable quantitative movement analysis using single-camera videos}.
\emph{Nature Communications} \textbf{11}, 1--10 (2020).}

\leavevmode\vadjust pre{\hypertarget{ref-stenum_two-dimensional_2021}{}}%
\CSLLeftMargin{14. }
\CSLRightInline{Stenum, J., Rossi, C. \& Roemmich, R. T.
\href{https://doi.org/10.1371/journal.pcbi.1008935}{Two-dimensional
video-based analysis of human gait using pose estimation}. \emph{PLOS
Computational Biology} \textbf{17}, e1008935 (2021).}

\leavevmode\vadjust pre{\hypertarget{ref-mehdizadeh_concurrent_2021}{}}%
\CSLLeftMargin{15. }
\CSLRightInline{Mehdizadeh, S. \emph{et al.}
\href{https://doi.org/10.1186/s12984-021-00933-0}{Concurrent validity of
human pose tracking in video for measuring gait parameters in older
adults: a preliminary analysis with multiple trackers, viewing angles,
and walking directions}. \emph{Journal of NeuroEngineering and
Rehabilitation} \textbf{18}, 139 (2021).}

\leavevmode\vadjust pre{\hypertarget{ref-Cao2016}{}}%
\CSLLeftMargin{16. }
\CSLRightInline{Cao, Z., Simon, T., Wei, S.-E. \& Sheikh, Y.
\href{http://arxiv.org/abs/1611.08050}{Realtime Multi-Person 2D Pose
Estimation using Part Affinity Fields}. (2016).}

\leavevmode\vadjust pre{\hypertarget{ref-Martinez2017}{}}%
\CSLLeftMargin{17. }
\CSLRightInline{Martinez, J., Hossain, R., Romero, J. \& Little, J. J.
\href{https://doi.org/10.1109/ICCV.2017.288}{A Simple Yet Effective
Baseline for 3d Human Pose Estimation}. \emph{Proceedings of the IEEE
International Conference on Computer Vision} \textbf{2017-Octob},
2659--2668 (2017).}

\leavevmode\vadjust pre{\hypertarget{ref-Pavllo2018}{}}%
\CSLLeftMargin{18. }
\CSLRightInline{Pavllo, D., Feichtenhofer, C., Grangier, D. \& Auli, M.
\href{http://arxiv.org/abs/1811.11742}{3D human pose estimation in video
with temporal convolutions and semi-supervised training}.
\emph{Proceedings of the IEEE Computer Society Conference on Computer
Vision and Pattern Recognition} \textbf{2019-June}, 7745--7754 (2018).}

\leavevmode\vadjust pre{\hypertarget{ref-Loper2015}{}}%
\CSLLeftMargin{19. }
\CSLRightInline{Loper, M., Mahmood, N., Romero, J., Pons-Moll, G. \&
Black, M. J. \href{https://doi.org/10.1145/2816795.2818013}{SMPL}.
\emph{ACM Transactions on Graphics} \textbf{34}, 1--16 (2015).}

\leavevmode\vadjust pre{\hypertarget{ref-kanazawa_end_2018}{}}%
\CSLLeftMargin{20. }
\CSLRightInline{Kanazawa, A., Black, M. J., Jacobs, D. W. \& Malik, J.
End-to-End Recovery of Human Shape and Pose. in \emph{2018 IEEE/CVF
Conference on Computer Vision and Pattern Recognition} 7122--7131
(2018).
doi:\href{https://doi.org/10.1109/CVPR.2018.00744}{10.1109/CVPR.2018.00744}.}

\leavevmode\vadjust pre{\hypertarget{ref-kolotouros_learning_2019}{}}%
\CSLLeftMargin{21. }
\CSLRightInline{Kolotouros, N., Pavlakos, G., Black, M. \& Daniilidis,
K. Learning to Reconstruct 3D Human Pose and Shape via Model-Fitting in
the Loop. in \emph{2019 IEEE/CVF International Conference on Computer
Vision (ICCV)} 2252--2261 (2019).
doi:\href{https://doi.org/10.1109/ICCV.2019.00234}{10.1109/ICCV.2019.00234}.}

\leavevmode\vadjust pre{\hypertarget{ref-kocabas_vibe_2020}{}}%
\CSLLeftMargin{22. }
\CSLRightInline{Kocabas, M., Athanasiou, N. \& Black, M. J.
\href{https://openaccess.thecvf.com/content_CVPR_2020/html/Kocabas_VIBE_Video_Inference_for_Human_Body_Pose_and_Shape_Estimation_CVPR_2020_paper.html}{VIBE:
Video Inference for Human Body Pose and Shape Estimation}. in
\emph{Proceedings of the IEEE/CVF Conference on Computer Vision and
Pattern Recognition} 5253--5263 (2020).}

\leavevmode\vadjust pre{\hypertarget{ref-liu_graph_2020}{}}%
\CSLLeftMargin{23. }
\CSLRightInline{Liu, J., Rojas, J., Liang, Z., Li, Y. \& Guan, Y.
\href{https://arxiv.org/abs/2003.14179v4}{A Graph Attention
Spatio-temporal Convolutional Network for 3D Human Pose Estimation in
Video}. (2020).}

\leavevmode\vadjust pre{\hypertarget{ref-Bogo2016}{}}%
\CSLLeftMargin{24. }
\CSLRightInline{Bogo, F. \emph{et al.} Keep It SMPL: Automatic
Estimation of 3D Human Pose and Shape from a Single Image. in \emph{ECCV
2016: Computer Vision -- ECCV 2016} 561--578 (Springer, Cham, 2016).
doi:\href{https://doi.org/10.1007/978-3-319-46454-1_34}{10.1007/978-3-319-46454-1\_34}.}

\leavevmode\vadjust pre{\hypertarget{ref-pavlakos_expressive_2019}{}}%
\CSLLeftMargin{25. }
\CSLRightInline{Pavlakos, G. \emph{et al.} Expressive Body Capture: 3D
Hands, Face, and Body From a Single Image. in \emph{2019 IEEE/CVF
Conference on Computer Vision and Pattern Recognition (CVPR)}
10967--10977 (IEEE, 2019).
doi:\href{https://doi.org/10.1109/CVPR.2019.01123}{10.1109/CVPR.2019.01123}.}

\leavevmode\vadjust pre{\hypertarget{ref-wu_isb_2002}{}}%
\CSLLeftMargin{26. }
\CSLRightInline{Wu, G. \emph{et al.}
\href{https://doi.org/10.1016/s0021-9290(01)00222-6}{ISB recommendation
on definitions of joint coordinate system of various joints for the
reporting of human joint motion-\/-part I: ankle, hip, and spine.
International Society of Biomechanics}. \emph{J Biomech} \textbf{35},
543--548 (2002).}

\leavevmode\vadjust pre{\hypertarget{ref-Wu2005}{}}%
\CSLLeftMargin{27. }
\CSLRightInline{Wu, G. \emph{et al.}
\href{https://www.ncbi.nlm.nih.gov/pubmed/15844264}{ISB recommendation
on definitions of joint coordinate systems of various joints for the
reporting of human joint motion-\/-Part II: shoulder, elbow, wrist and
hand.} \emph{Journal of biomechanics} \textbf{38}, 981--992 (2005).}

\leavevmode\vadjust pre{\hypertarget{ref-Cotton2020}{}}%
\CSLLeftMargin{28. }
\CSLRightInline{Cotton, R. J.
\href{https://www.computer.org/csdl/proceedings-article/fg/2020/307900a588/1kecIJHALKM}{Kinematic
Tracking of Rehabilitation Patients With Markerless Pose Estimation
Fused with Wearable Inertial Sensors}. \emph{IEEE 15th International
Conference on Automatic Face \& Gesture Recognition} (2020).}

\leavevmode\vadjust pre{\hypertarget{ref-yatsenko_datajoint_2015}{}}%
\CSLLeftMargin{29. }
\CSLRightInline{Yatsenko, D. \emph{et al.} \emph{DataJoint: managing big
scientific data using MATLAB or Python}. 031658
\url{https://www.biorxiv.org/content/10.1101/031658v1} (2015)
doi:\href{https://doi.org/10.1101/031658}{10.1101/031658}.}

\leavevmode\vadjust pre{\hypertarget{ref-yatsenko_datajoint_2018}{}}%
\CSLLeftMargin{30. }
\CSLRightInline{Yatsenko, D., Walker, E. Y. \& Tolias, A. S.
\href{http://arxiv.org/abs/1807.11104}{DataJoint: A Simpler Relational
Data Model}. \emph{arXiv:1807.11104 {[}cs{]}} (2018).}

\leavevmode\vadjust pre{\hypertarget{ref-datajoint_team_datajoint_2022}{}}%
\CSLLeftMargin{31. }
\CSLRightInline{DataJoint Team. DataJoint Projects.
\url{https://www.datajoint.com/projects} (2022).}

\leavevmode\vadjust pre{\hypertarget{ref-mmtracking_contributors_mmtracking_2020}{}}%
\CSLLeftMargin{32. }
\CSLRightInline{MMTracking Contributors.
\emph{\href{https://github.com/open-mmlab/mmtracking}{MMTracking:
OpenMMLab}}. (2020).}

\leavevmode\vadjust pre{\hypertarget{ref-wojke_simple_2017}{}}%
\CSLLeftMargin{33. }
\CSLRightInline{Wojke, N., Bewley, A. \& Paulus, D.
\href{http://arxiv.org/abs/1703.07402}{Simple Online and Realtime
Tracking with a Deep Association Metric}. \emph{arXiv:1703.07402
{[}cs{]}} (2017).}

\leavevmode\vadjust pre{\hypertarget{ref-zhang_fairmot_2020}{}}%
\CSLLeftMargin{34. }
\CSLRightInline{Zhang, Y., Wang, C., Wang, X., Zeng, W. \& Liu, W.
\href{http://arxiv.org/abs/2004.01888}{FairMOT: On the Fairness of
Detection and Re-Identification in Multiple Object Tracking}.
\emph{arXiv:2004.01888 {[}cs{]}} (2020).}

\leavevmode\vadjust pre{\hypertarget{ref-wu_track_2021}{}}%
\CSLLeftMargin{35. }
\CSLRightInline{Wu, J. \emph{et al.} Track to Detect and Segment: An
Online Multi-Object Tracker. in \emph{2021 IEEE/CVF Conference on
Computer Vision and Pattern Recognition (CVPR)} 12347--12356 (2021).
doi:\href{https://doi.org/10.1109/CVPR46437.2021.01217}{10.1109/CVPR46437.2021.01217}.}

\leavevmode\vadjust pre{\hypertarget{ref-Sun2020}{}}%
\CSLLeftMargin{36. }
\CSLRightInline{Sun, P. \emph{et al.}
\href{http://arxiv.org/abs/2012.15460}{TransTrack: Multiple-Object
Tracking with Transformer}. (2020).}

\leavevmode\vadjust pre{\hypertarget{ref-mmpose_contributors_openmmlab_2020}{}}%
\CSLLeftMargin{37. }
\CSLRightInline{MMPose Contributors.
\emph{\href{https://github.com/open-mmlab/mmpose}{OpenMMLab Pose
Estimation Toolbox and Benchmark}}. (2020).}

\leavevmode\vadjust pre{\hypertarget{ref-kocabas_spec_2021}{}}%
\CSLLeftMargin{38. }
\CSLRightInline{Kocabas, M. \emph{et al.}
\href{http://arxiv.org/abs/2110.00620}{SPEC: Seeing People in the Wild
with an Estimated Camera}. \emph{arXiv:2110.00620 {[}cs{]}} (2021).}

\leavevmode\vadjust pre{\hypertarget{ref-kocabas_pare_2021}{}}%
\CSLLeftMargin{39. }
\CSLRightInline{Kocabas, M., Huang, C.-H. P., Hilliges, O. \& Black, M.
J. \href{http://arxiv.org/abs/2104.08527}{PARE: Part Attention Regressor
for 3D Human Body Estimation}. \emph{arXiv:2104.08527 {[}cs{]}} (2021).}

\leavevmode\vadjust pre{\hypertarget{ref-vedaldi_monocular_2020}{}}%
\CSLLeftMargin{40. }
\CSLRightInline{Choutas, V., Pavlakos, G., Bolkart, T., Tzionas, D. \&
Black, M. J.
\href{https://doi.org/10.1007/978-3-030-58607-2_2}{Monocular Expressive
Body Regression Through Body-Driven Attention}. in \emph{Computer Vision
-- ECCV 2020} (eds. Vedaldi, A., Bischof, H., Brox, T. \& Frahm, J.-M.)
vol. 12355 20--40 (Springer International Publishing, 2020).}

\leavevmode\vadjust pre{\hypertarget{ref-feng_collaborative_2021}{}}%
\CSLLeftMargin{41. }
\CSLRightInline{Feng, Y., Choutas, V., Bolkart, T., Tzionas, D. \&
Black, M. J. \href{http://arxiv.org/abs/2105.05301}{Collaborative
Regression of Expressive Bodies using Moderation}.
\emph{arXiv:2105.05301 {[}cs{]}} (2021).}

\leavevmode\vadjust pre{\hypertarget{ref-kolotouros_probabilistic_2021}{}}%
\CSLLeftMargin{42. }
\CSLRightInline{Kolotouros, N., Pavlakos, G., Jayaraman, D. \&
Daniilidis, K. \href{http://arxiv.org/abs/2108.11944}{Probabilistic
Modeling for Human Mesh Recovery}. \emph{arXiv:2108.11944 {[}cs{]}}
(2021).}

\leavevmode\vadjust pre{\hypertarget{ref-rempe_humor_2021}{}}%
\CSLLeftMargin{43. }
\CSLRightInline{Rempe, D. \emph{et al.} HuMoR: 3D Human Motion Model for
Robust Pose Estimation. in \emph{International Conference on Computer
Vision (ICCV)} (2021).}

\leavevmode\vadjust pre{\hypertarget{ref-Sun2019}{}}%
\CSLLeftMargin{44. }
\CSLRightInline{Sun, K., Xiao, B., Liu, D. \& Wang, J.
\href{https://doi.org/10.1109/CVPR.2019.00584}{Deep High-Resolution
Representation Learning for Human Pose Estimation}. \emph{Proceedings of
the IEEE Computer Society Conference on Computer Vision and Pattern
Recognition} \textbf{2019-June}, 5686--5696 (2019).}

\leavevmode\vadjust pre{\hypertarget{ref-zhang_distribution-aware_2020}{}}%
\CSLLeftMargin{45. }
\CSLRightInline{Zhang, F., Zhu, X., Dai, H., Ye, M. \& Zhu, C.
Distribution-Aware Coordinate Representation for Human Pose Estimation.
in \emph{2020 IEEE/CVF Conference on Computer Vision and Pattern
Recognition (CVPR)} 7091--7100 (2020).
doi:\href{https://doi.org/10.1109/CVPR42600.2020.00712}{10.1109/CVPR42600.2020.00712}.}

\leavevmode\vadjust pre{\hypertarget{ref-lin_microsoft_2014}{}}%
\CSLLeftMargin{46. }
\CSLRightInline{Lin, T.-Y. \emph{et al.} Microsoft COCO: Common Objects
in Context. in \emph{Computer Vision -- ECCV 2014} (eds. Fleet, D.,
Pajdla, T., Schiele, B. \& Tuytelaars, T.) 740--755 (Springer
International Publishing, 2014).
doi:\href{https://doi.org/10.1007/978-3-319-10602-1_48}{10.1007/978-3-319-10602-1\_48}.}

\leavevmode\vadjust pre{\hypertarget{ref-jin_whole-body_2020}{}}%
\CSLLeftMargin{47. }
\CSLRightInline{Jin, S. \emph{et al.} Whole-Body Human Pose Estimation
in the Wild. in \emph{Computer Vision -- ECCV 2020} (eds. Vedaldi, A.,
Bischof, H., Brox, T. \& Frahm, J.-M.) 196--214 (Springer International
Publishing, 2020).
doi:\href{https://doi.org/10.1007/978-3-030-58545-7_12}{10.1007/978-3-030-58545-7\_12}.}

\leavevmode\vadjust pre{\hypertarget{ref-Mathis2018}{}}%
\CSLLeftMargin{48. }
\CSLRightInline{Mathis, A. \emph{et al.} DeepLabCut: markerless pose
estimation of user-defined body parts with deep learning. \emph{Nature
Neuroscience 2018} 1 (2018)
doi:\href{https://doi.org/10.1038/s41593-018-0209-y}{10.1038/s41593-018-0209-y}.}

\leavevmode\vadjust pre{\hypertarget{ref-Nath2019}{}}%
\CSLLeftMargin{49. }
\CSLRightInline{Nath, T. \emph{et al.} Using DeepLabCut for 3D
markerless pose estimation across species and behaviors. \emph{Nature
Protocols} 1 (2019)
doi:\href{https://doi.org/10.1038/s41596-019-0176-0}{10.1038/s41596-019-0176-0}.}

\leavevmode\vadjust pre{\hypertarget{ref-rajasegaran_tracking_2021}{}}%
\CSLLeftMargin{50. }
\CSLRightInline{Rajasegaran, J., Pavlakos, G., Kanazawa, A. \& Malik, J.
\href{http://arxiv.org/abs/2111.07868}{Tracking People with 3D
Representations}. \emph{arXiv:2111.07868 {[}cs{]}} (2021).}

\leavevmode\vadjust pre{\hypertarget{ref-yuan_glamr_2021}{}}%
\CSLLeftMargin{51. }
\CSLRightInline{Yuan, Y., Iqbal, U., Molchanov, P., Kitani, K. \& Kautz,
J. \href{https://arxiv.org/abs/2112.01524v1}{GLAMR: Global
Occlusion-Aware Human Mesh Recovery with Dynamic Cameras}. (2021).}

\leavevmode\vadjust pre{\hypertarget{ref-lonini_video-based_2022}{}}%
\CSLLeftMargin{52. }
\CSLRightInline{Lonini, L. \emph{et al.}
\href{https://doi.org/10.1159/000520732}{Video-Based Pose Estimation for
Gait Analysis in Stroke Survivors during Clinical Assessments: A
Proof-of-Concept Study}. \emph{DIB} \textbf{6}, 9--18 (2022).}

\leavevmode\vadjust pre{\hypertarget{ref-holzinger_causability_2019}{}}%
\CSLLeftMargin{53. }
\CSLRightInline{Holzinger, A., Langs, G., Denk, H., Zatloukal, K. \&
MÃŒller, H. \href{https://doi.org/10.1002/widm.1312}{Causability and
explainability of artificial intelligence in medicine}. \emph{Wiley
Interdiscip Rev Data Min Knowl Discov} \textbf{9}, e1312 (2019).}

\leavevmode\vadjust pre{\hypertarget{ref-amann_explainability_2020}{}}%
\CSLLeftMargin{54. }
\CSLRightInline{Amann, J. \emph{et al.}
\href{https://doi.org/10.1186/s12911-020-01332-6}{Explainability for
artificial intelligence in healthcare: a multidisciplinary perspective}.
\emph{BMC Medical Informatics and Decision Making} \textbf{20}, 310
(2020).}

\leavevmode\vadjust pre{\hypertarget{ref-vilone_notions_2021}{}}%
\CSLLeftMargin{55. }
\CSLRightInline{Vilone, G. \& Longo, L.
\href{https://doi.org/10.1016/j.inffus.2021.05.009}{Notions of
explainability and evaluation approaches for explainable artificial
intelligence}. \emph{Information Fusion} \textbf{76}, 89--106 (2021).}

\leavevmode\vadjust pre{\hypertarget{ref-lu_quantifying_2021}{}}%
\CSLLeftMargin{56. }
\CSLRightInline{Lu, M. \emph{et al.}
\href{https://doi.org/10.1016/j.media.2021.102179}{Quantifying
Parkinson's disease motor severity under uncertainty using MDS-UPDRS
videos}. \emph{Medical Image Analysis} \textbf{73}, 102179 (2021).}

\leavevmode\vadjust pre{\hypertarget{ref-mehrabi_survey_2021}{}}%
\CSLLeftMargin{57. }
\CSLRightInline{Mehrabi, N., Morstatter, F., Saxena, N., Lerman, K. \&
Galstyan, A. \href{https://doi.org/10.1145/3457607}{A Survey on Bias and
Fairness in Machine Learning}. \emph{ACM Comput. Surv.} \textbf{54},
115:1--115:35 (2021).}

\leavevmode\vadjust pre{\hypertarget{ref-pessach_algorithmic_2020}{}}%
\CSLLeftMargin{58. }
\CSLRightInline{Pessach, D. \& Shmueli, E.
\href{http://arxiv.org/abs/2001.09784}{Algorithmic Fairness}.
\emph{arXiv:2001.09784 {[}cs, stat{]}} (2020).}

\leavevmode\vadjust pre{\hypertarget{ref-Chouldechova2018}{}}%
\CSLLeftMargin{59. }
\CSLRightInline{Chouldechova, A. \& Roth, A. \emph{The Frontiers of
Fairness in Machine Learning}. (2018).}

\leavevmode\vadjust pre{\hypertarget{ref-trewin_considerations_2019}{}}%
\CSLLeftMargin{60. }
\CSLRightInline{Trewin, S. \emph{et al.}
\href{https://doi.org/10.1145/3362077.3362086}{Considerations for AI
fairness for people with disabilities}. \emph{AI Matters} \textbf{5},
40--63 (2019).}

\leavevmode\vadjust pre{\hypertarget{ref-the_pandas_development_team_pandas-devpandas_2020}{}}%
\CSLLeftMargin{61. }
\CSLRightInline{The pandas development team. \emph{pandas-dev/pandas:
Pandas 1.0.3}. (Zenodo, 2020).
doi:\href{https://doi.org/10.5281/zenodo.3715232}{10.5281/zenodo.3715232}.}

\leavevmode\vadjust pre{\hypertarget{ref-yatsenko_datajoint_2021}{}}%
\CSLLeftMargin{62. }
\CSLRightInline{Yatsenko, D. \emph{et al.} \emph{DataJoint Elements:
Data Workflows for Neurophysiology}. 2021.03.30.437358
\url{https://www.biorxiv.org/content/10.1101/2021.03.30.437358v2} (2021)
doi:\href{https://doi.org/10.1101/2021.03.30.437358}{10.1101/2021.03.30.437358}.}

\leavevmode\vadjust pre{\hypertarget{ref-yuan_simpoe_2021}{}}%
\CSLLeftMargin{63. }
\CSLRightInline{Yuan, Y., Wei, S.-E., Simon, T., Kitani, K. \& Saragih,
J. SimPoE: Simulated Character Control for 3D Human Pose Estimation. in
\emph{2021 IEEE/CVF Conference on Computer Vision and Pattern
Recognition (CVPR)} 7155--7165 (2021).
doi:\href{https://doi.org/10.1109/CVPR46437.2021.00708}{10.1109/CVPR46437.2021.00708}.}

\leavevmode\vadjust pre{\hypertarget{ref-Shi2020}{}}%
\CSLLeftMargin{64. }
\CSLRightInline{Shi, M. \emph{et al.}
\href{http://arxiv.org/abs/2006.12075}{MotioNet: 3D Human Motion
Reconstruction from Monocular Video with Skeleton Consistency}.
\emph{ACM Transactions on Graphics} \textbf{40}, (2020).}

\leavevmode\vadjust pre{\hypertarget{ref-shimada_neural_2021}{}}%
\CSLLeftMargin{65. }
\CSLRightInline{Shimada, S., Golyanik, V., Xu, W., PÃ©rez, P. \&
Theobalt, C. \href{http://arxiv.org/abs/2105.01057}{Neural Monocular 3D
Human Motion Capture with Physical Awareness}. \emph{arXiv:2105.01057
{[}cs{]}} (2021).}

\leavevmode\vadjust pre{\hypertarget{ref-shimada_physcap_2020}{}}%
\CSLLeftMargin{66. }
\CSLRightInline{Shimada, S., Golyanik, V., Xu, W. \& Theobalt, C.
\href{http://arxiv.org/abs/2008.08880}{PhysCap: Physically Plausible
Monocular 3D Motion Capture in Real Time}. \emph{arXiv:2008.08880
{[}cs{]}} (2020).}

\leavevmode\vadjust pre{\hypertarget{ref-gu_ava_2018}{}}%
\CSLLeftMargin{67. }
\CSLRightInline{Gu, C. \emph{et al.} AVA: A Video Dataset of
Spatio-Temporally Localized Atomic Visual Actions. in \emph{2018
IEEE/CVF Conference on Computer Vision and Pattern Recognition}
6047--6056 (2018).
doi:\href{https://doi.org/10.1109/CVPR.2018.00633}{10.1109/CVPR.2018.00633}.}

\leavevmode\vadjust pre{\hypertarget{ref-Cotton2019}{}}%
\CSLLeftMargin{68. }
\CSLRightInline{Cotton, R. J. \& Rogers, J.
\href{https://doi.org/10.1109/ICORR.2019.8779538}{Wearable Monitoring of
Joint Angle and Muscle Activity}. in \emph{2019 IEEE 16th International
Conference on Rehabilitation Robotics (ICORR)} vol. 2019 258--263 (IEEE,
2019).}

\leavevmode\vadjust pre{\hypertarget{ref-richard_whittles_2012}{}}%
\CSLLeftMargin{69. }
\CSLRightInline{Richard, J., Levine, D. \& Whittle, M.
\emph{\href{https://www.elsevier.com/books/whittles-gait-analysis/levine/978-0-7020-4265-2}{Whittle's
Gait Analysis - 5th Edition}}. (Elsevier, 2012).}

\leavevmode\vadjust pre{\hypertarget{ref-whittle_clinical_1996}{}}%
\CSLLeftMargin{70. }
\CSLRightInline{Whittle, M. W.
\href{https://doi.org/10.1016/0167-9457(96)00006-1}{Clinical gait
analysis: A review}. \emph{Human Movement Science} \textbf{15}, 369--387
(1996).}

\end{CSLReferences}

\end{document}